\documentclass[a4paper,fleqn]{cas-dc}

\usepackage[numbers]{natbib}

\DeclareUnicodeCharacter{2212}{-}
\usepackage{times}
\usepackage{epsfig}
\usepackage{graphicx}
\usepackage{amsmath}
\usepackage{amssymb}
\usepackage{bm}
\usepackage{color}
\usepackage{ulem}
\usepackage{diagbox}

\usepackage{amsthm}
\usepackage{algorithmic}
\usepackage{algorithm}
\usepackage{caption}

\definecolor{fuchsia}{rgb}{0.79, 0.17, 0.57}
\newcommand{\red}[1]{\textbf{{\color{red} #1}}}
\newcommand{\gray}[1]{\textbf{{\color{gray} #1}}}
\newcommand{\orange}[1]{\textbf{{\color{orange} #1}}}
\newcommand{\cyan}[1]{\textbf{{\color{cyan} #1}}}
\newcommand{\fuchsia}[1]{\textbf{{\color{fuchsia} #1}}}

\newbox\jsavebox
\newcommand{\jsubfig}[2]{%
	\sbox\jsavebox{#1}%
	\parbox[t]{\wd\jsavebox}{\centering\usebox\jsavebox\\#2}%
	}

\usepackage{mathtools}

\DeclarePairedDelimiterX{\infdivx}[2]{(}{)}{%
	#1\;\delimsize\|\;#2%
}

\usepackage{xcolor}

\newcommand{\ra}[1]{\renewcommand{\arraystretch}{#1}}

\usepackage{subcaption}

\begin{document}
\let\WriteBookmarks\relax
\def\floatpagepagefraction{1}
\def\textpagefraction{.001}
\shorttitle{Implicit Pairs for Boosting Unpaired Image-to-Image Translation}
\shortauthors{Y. Ginger et~al.}

\title [mode = title]{Implicit Pairs for Boosting Unpaired Image-to-Image Translation}                      

\address[]{Tel Aviv University}
\cortext[cor1]{Corresponding author}

\author[]{Yiftach Ginger}[orcid=0000-0001-8575-7405]
\ead{iftachg@mail.tau.ac.il}
\cormark[1]

\author[]{Dov Danon}

\author[]{Hadar Averbuch-Elor}

\author[]{Daniel Cohen-Or}


\begin{abstract}
In image-to-image translation the goal is to learn a mapping from one image domain to another. In the case of supervised  approaches the  mapping is learned from  paired samples. However, collecting large sets of image pairs is often either prohibitively expensive or not possible.
As a result, in recent years more attention has been given to techniques that learn the mapping from unpaired sets.

In our work, we show that injecting implicit pairs into unpaired sets strengthens the mapping between the two domains, improves the compatibility of their distributions, and leads to performance boosting of unsupervised techniques by up to 12\% across several measurements. 

The competence of the implicit pairs is further displayed with the use of pseudo-pairs, i.e., paired samples which only approximate a real pair. We  demonstrate  the  effect  of  the  approximated implicit samples on image-to-image translation problems, where such pseudo-pairs may be synthesized in one direction, but not in the other. We further show that pseudo-pairs are significantly more effective as implicit pairs in an unpaired setting, than directly using them explicitly in a paired setting.
\end{abstract}


\maketitle

\section{Introduction}

The goal of image-to-image  translation is to learn a mapping from one image domain to another. In recent years, a plethora of methods has arisen to solve the problem using deep neural networks. A straightforward supervised approach is to learn the mapping from paired samples \cite{DBLP:journals/corr/IsolaZZE16}. However, collecting large sets of image pairs is often prohibitively expensive or infeasible. Learning the mapping from unpaired data is thus more attractive, but significantly more technically challenging, as the problem becomes highly under-constrained. Many solutions were suggested which find proxies for the signal of paired samples \cite{DBLP:journals/corr/ZhuPIE17, DBLP:journals/corr/YiZTG17, DBLP:journals/corr/KimCKLK17}, but ultimately they still reach worse performance than equivalent supervised versions. Our aim is to improve these results without requiring costly collection of supervised data by improving and augmenting the data which the models learn from.

Data augmentation is a well known and widely used method to improve learning processes by augmenting the distribution of the training data with new samples and has been using for training neural networks since at least lenet-5 \cite{Lecun98gradient}. Common and widely used augmentation techniques are geometric transformation such as flipping, rotating, cropping and translating the images \cite{DA_Survey}.

Using data augmentation methods is also common practice when training GANs. Most common are random flipping and cropping of the data \cite{DBLP:journals/corr/ZhuPIE17, DBLP:journals/corr/IsolaZZE16, U-GAT-IT, TravelGAN, DBLP:journals/corr/LiuBK17} but other, less prevalent practices are used such as random jittering \cite{DBLP:journals/corr/IsolaZZE16}, color space translation and grayscale inversion \cite{DBLP:journals/corr/LiuBK17}.
Some case-specific augmentations are crafted \cite{U-NET, Fang2019InstaBoostBI, DBLP:journals/corr/MilletariNA16, DBLP:journals/corr/RothLSSKYLS15} or learned \cite{DBLP:journals/corr/HaubergFLFH15, DBLP:journals/corr/abs-1805-09501, Ratner2017LearningTC} to complement the specific task  while others aim at utilizing known theoretical and empiric qualities of the algorithms which they are used to improve \cite{DBLP:journals/corr/abs-1708-04552, DBLP:journals/corr/abs-1906-06423, DBLP:journals/corr/WangXW015, DBLP:journals/corr/abs-1904-12848}.
However, although augmentation methods have been shown to be of great benefit when training models, to the best of our knowledge there has been no research into data augmentation methods which leverage the peculiarities of the GAN framework and our understanding of it.

In this paper we propose an augmentation method specific for the unsupervised Image-to-Image translation framework in which synthetic samples are injected to the datasets to form pairings.
We start by establishing our main claim that Image-to-Image translation models are able to utilize pairing information even in unsupervised training regimes and demonstrate its veracity with extensive experimental results. We then show that the quality of the mappings learned depends on the portion of implicit pairs in the dataset. Following that, we analyze these results and establish an interpretation for how an unsupervised model could reliably benefit from implicit pairs. 

This non-intuitive finding encourages the use of pseudo-pairs in an unsupervised setting. We propose this as a data augmentation method in which synthetic samples are generated to construct pseudo-pairs and enrich the datasets. We further detail our approach, sample synthesis methods and our evaluation metrics.

Finally, we demonstrate the efficacy of pseudo-pairs for solving unsupervised Image-to-Image translation problems. We further argue that pseudo-pairs are significantly more effective when used as implicit pairs in an unpaired setting, than when used explicitly in a paired setting.

Explicitly stated, our contributions are:
\begin{itemize}
	
	\item We demonstrate that unsupervised image-to-image translation networks benefit from the latent signal that pairs add to the dataset.
	
	\item We analyze the effect of the  percentage of pairs in the dataset and demonstrate that having even a small percentage of pairs enhances the datasets and allows the model to reach peak performance rates.
	
	\item We introduce a unique data augmentation method for the image-to-image translation framework and demonstrate that it is more effective in an unpaired setting than as explicit pairs in a paired setting.
	
\end{itemize}

\section{Related Works}

Pix2Pix \cite{DBLP:journals/corr/IsolaZZE16} was the first successful attempt to use a conditional GAN to learn a mapping between two image distributions. As a supervised method it requires paired samples, one from each distribution, to be explicitly linked in the training phase.


Since gathering a large paired dataset can be difficult and expensive, various unsupervised architectures were suggested which do not require such explicit pairing \cite{DBLP:journals/corr/ZhuPIE17,DBLP:journals/corr/LiuBK17,DBLP:journals/corr/YiZTG17,DBLP:journals/corr/KimCKLK17,DBLP:journals/corr/BenaimW17, U-GAT-IT, TravelGAN}.
Several methods bridge the difference between supervised and unsupervised architectures by allowing the use of a small set of paired images, together with a large set of unpaired ones in a semi-supervised fashion.
They accomplish this by alternating between supervised and unsupervised phases during training \cite{DBLP:journals/corr/abs-1805-10790,DBLP:journals/corr/abs-1805-03189}. Other semi-supervised solutions separate the learning of the joint distribution and the marginal distribution of the domains, by independently learning the translation from the supervised set and the unsupervised set \cite{DBLP:journals/corr/abs-1709-06548, DBLP:journals/corr/LiXZZ17}. 

\medskip

Deep neural networks require large amounts of data to train properly, which can prove prohibitively expensive in some cases.
To cope with this problem various methods have been devised to create more samples by augmenting existing data into new samples in order to create meaningful expressions of the underlying distribution. Simple augmentation methods for images include rotation, skewing, cropping and other affine transformations. These simple methods are quite ubiquitous and have been used since the early days of deep neural networks \cite{Lecun98gradient}, but are limited in the amount of data they can generate, as well as the amount of effective information that they add to the dataset.

Other more complex augmentation methods could be model-based, use learned generative models, and even GANs \cite{10.1007/978-3-319-66182-7_26,10.1007/978-3-319-46073-4_2,TUSTISON2018,2017arXiv171104340A,DBLP:journals/corr/abs-1801-05401,DBLP:journals/corr/abs-1803-01229,2018arXiv180309655M,2016arXiv161101331S}.

Lastly, while there are works which tailor on augmentation methods for specific algorithms or tasks \cite{DBLP:journals/corr/abs-1904-12848, DBLP:journals/corr/WangXW015, DBLP:journals/corr/abs-1906-06423, DBLP:journals/corr/abs-1708-04552, Fang2019InstaBoostBI, DBLP:journals/corr/RothLSSKYLS15, DBLP:journals/corr/MilletariNA16, U-NET} we are not aware of any such prior work that is focused on augmenting a dataset that is used to train a GAN. This leaves the field restricted in ways that other ML domains are not.

\section{Implicit pairs}
\label{sec:implicit_pairs}

Denote $T: A \rightarrow B$ as a translation we wish to learn. A \textit{pair} are two samples $a \in A, b \in B$ s.t. $T(a) = b$ and a \textit{paired dataset} is a dataset $D = (A_D, B_D)$ where $\forall a \in A_D, \; \; T(a) \in B_D$

In the Image-to-Image translation literature pairs are mostly considered when used in a supervised manner, while unsupervised approaches do not consider whether there are any pairs in the dataset at all \cite{DBLP:journals/corr/ZhuPIE17, DBLP:journals/corr/LiuBK17, DBLP:journals/corr/YiZTG17, DBLP:journals/corr/KimCKLK17} unless they are used to augment the learning process in a semi-supervised model \cite{DBLP:journals/corr/abs-1805-10790, DBLP:journals/corr/abs-1805-03189}. This is done even though many unsupervised Image-to-Image translation papers show results on paired datasets and without considering what effect having paired samples in the dataset has on the learning process.

It seems reasonable not to consider the effect of paired samples on unsupervised learning as there is no explicit usage of such information in the algorithms. Consider for example the supervised Pix2Pix algorithm \cite{DBLP:journals/corr/IsolaZZE16} where samples are drawn in pairs $(a_i, T(a_i)$ and a translated image $G(a_i))$ is explicitly compared to the target paired sample $T(a_i)$ in the objective function. Its unsupervised variant, the CycleGAN algorithm \cite{DBLP:journals/corr/ZhuPIE17}, samples without regard to their pairing $(a_i,T(a_j))$ and does not even use the sample from domain $B$ when training the generator $G_A: A \rightarrow B$. Instead the domain information is only conferred through the discriminator which is exposed to $T(a_j)$, but is not explicitly expected to use any information about paired samples.

Consider two scenarios - in the first one we train an unsupervised Image-to-Image translation model using a paired dataset while in the second we only have an unpaired dataset, i.e. $\forall a \in A_D, \;  T(a) \notin B_D$. For simplicity assume that $A_D$ is shared between the scenarios.
If indeed unsupervised translation algorithms do not use the information inherent in the pairings we would expect similar results when training with either the paired or unpaired datasets. 

To evaluate this assumption we have conducted experiments where we train models using the exact same architecture, parameters and dataset size while varying the ratio of paired samples (denoted $\alpha$) in the dataset. Concretely, we train a dual generator-discriminator architecture (CycleGAN) on the following datasets: Cityscapes \cite{Cordts2016Cityscapes}, Facades \cite{Tylecek13} and CVC-14 \cite{s16060820}. We  split the datasets into train and test sets and sample the train sets to generate various $\alpha$-paired dataset configurations. In all our experiments, we select $|A| = |B|$ samples to generate balanced datasets. 

To evaluate the performance on the test set, we measure the MSE between the generated images and their true counterparts. Additionally we use the FCN-score metric introduced in \cite{DBLP:journals/corr/IsolaZZE16} to evaluate the learned translations for the Cityscapes \cite{Cordts2016Cityscapes} dataset. Please refer to the supplementary material for more information regarding the evaluation metric and its use as well as additional information regarding the CycleGAN architecture and parameters used.

\begin{table}

\centering
\ra{1.0}
\setlength{\tabcolsep}{2.35pt}
\begin{tabular}{l cccccccccc}
\toprule{} 
\phantom{} & \phantom{} & \phantom{} & \multicolumn{2}{c}{$Cityscapes$} & \phantom{}& \multicolumn{2}{c}{$CVC-14$} & \phantom{} & \multicolumn{2}{c}{$Facades$}  
\\

$\alpha$
      && & A2B            & B2A             &&  A2B           & B2A            && A2B            & B2A  \\
      \midrule{}
 0  (unpaired)  && & 0.26          & 0.22           && 0.23          & 0.24          && 0.36          & 0.84 \\  
 0.25 && & \textbf{0.24} & \textbf{0.21}  && 0.28          & 0.29          && \textbf{0.33} & 0.84 \\ 
 0.5  && & \textbf{0.24} & 0.22           && \textbf{0.22} & 0.23          && 0.37          & \textbf{0.80}  \\
 0.75 && & 0.27          & 0.22           && 0.24          & \textbf{0.22} && 0.37          & 0.84  \\
 1 (paired)   && & 0.25          & 0.22           && 0.23          & 0.25          && \textbf{0.33} & 0.87 \\    

\bottomrule{}
\end{tabular}
\caption{Reconstruction loss for different implicit pairing ratios, $\alpha$. lower is better. A2B is photo $\rightarrow$ labels, B2A is labels $\rightarrow$ photo.}
\label{table:first_stage_results}
\end{table}




\begin{table}
\centering
\begin{tabular}{ p{1.4cm} p{1.0cm} p{1.0cm} p{1.0cm} } 
$\alpha$ & 
 \multicolumn{1}{c}{\textbf{Per-pixel acc.}} & \multicolumn{1}{c}{\textbf{Per-class acc.}}
 & \multicolumn{1}{c}{\textbf{Class IOU}} \\
\hline
0 & 0.507 & 0.160 & 0.110 \\
0.25 & \textbf{0.566} & 0.162 & 0.111 \\
0.5 & 0.535 & \textbf{0.167} & 0.114 \\
0.75 & 0.542 & \textbf{0.167} & \textbf{0.118} \\
1 & 0.522 & 0.162 & 0.111
 
\end{tabular}
\caption{FCN-scores for different implicit pairing ratios, $\alpha$, on Cityscapes labels$\rightarrow$photo, higher is better. Remarkably using a mix of paired and unpaired samples is always better.}
\label{table:labels2photos_fcn8}
\end{table}
\begin{table}
\centering
\begin{tabular}{ p{1.4cm} p{1.0cm} p{1.0cm} p{1.0cm} } 
\textbf{$\alpha$} & 
 \multicolumn{1}{c}{\textbf{Per-pixel acc.}} & \multicolumn{1}{c}{\textbf{Per-class acc.}}
 & \multicolumn{1}{c}{\textbf{Class IOU}} \\
\hline
 0   & 0.582 & 0.212 & 0.158 \\
0.25 & 0.583 & 0.212 & 0.159 \\
0.5  & \textbf{0.598} & \textbf{0.221} & \textbf{0.166} \\
0.75 & 0.587 & 0.214 & 0.161 \\
1    & 0.582 & 0.208 & 0.156
 
\end{tabular}
\caption{FCN-scores for different implicit pairing ratios, $\alpha$, on Cityscapes photo$\rightarrow$labels, higher is better. Remarkably using a mix of paired and unpaired samples is always better.}
\label{table:photos2labels_fcn8}
\end{table}

We report our evaluation in Tables \ref{table:first_stage_results}, \ref{table:labels2photos_fcn8}, \ref{table:photos2labels_fcn8}. The first thing to note in the results is that $1$-paired datasets generally yield better performance than $0$-paired datasets with an improvement of up to 12\%. This shows that unsupervised Image-to-Image translation algorithms such as CycleGAN do indeed use implicit pairing information. Even more interesting is the fact that having even a low $25\%$ of the samples paired improves the results dramatically compared to having no pairs at all. Remarkably and unexpectedly, it seems that in most cases using a completely paired dataset is not the best option. Instead using a mix of paired and unpaired samples is usually a better strategy, surpassing completely paired datasets on average by 3.4\%. 

\begin{figure}
\centering%
\vspace{7pt}

\jsubfig{\includegraphics[height=3.6cm]{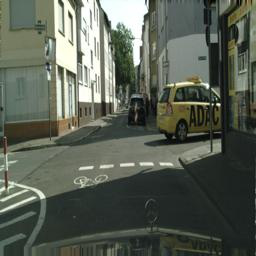}}
{}%
\hspace{0.2cm}
\jsubfig{\includegraphics[height=3.6cm]{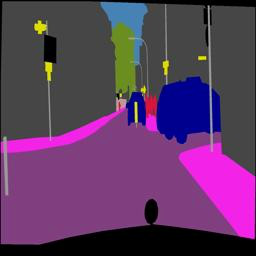}}
{}%
\\
	\jsubfig{\includegraphics[height=3.6cm]{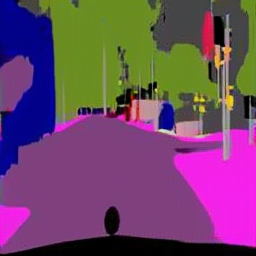}}
	{\vspace{-13pt}\colorbox{blue!30}{\small{0.046}}}%
\hspace{0.2cm}
	\jsubfig{\includegraphics[height=3.6cm]{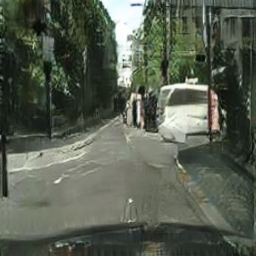}}
	{\vspace{-13pt}\colorbox{blue!30}{\small{0.485}}}%
\\
	\jsubfig{\includegraphics[height=3.6cm]{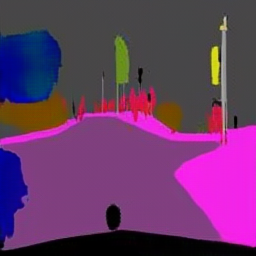}}
	{\vspace{-13pt}\colorbox{blue!30}{\small{0.088}}}%
\hspace{0.2cm}
	\jsubfig{\includegraphics[height=3.6cm]{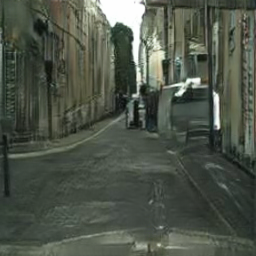}}
	{\vspace{-13pt}\colorbox{blue!30}{\small{0.686}}}%
\\

	\jsubfig{\includegraphics[height=3.6cm]{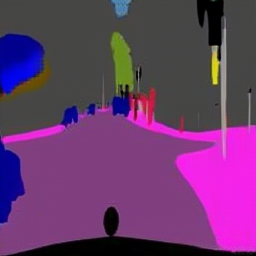}}
	{\vspace{-13pt}\colorbox{blue!30}{\small{0.090}}}%
\hspace{0.2cm}
	\jsubfig{\includegraphics[height=3.6cm]{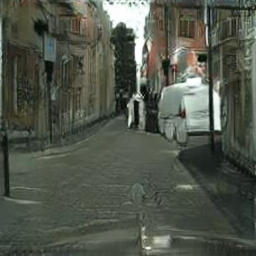}}
	{\vspace{-13pt}\colorbox{blue!30}{\small{0.712} }}%

\vspace{3pt}
\caption{\textbf{Illustration of implicit pairs ratio experiments.} Pixel accuracy for a random test sample using models trained with varying pairing ratios. From top to bottom: Source image, $\alpha=0$, $\alpha=0.5$, $\alpha=1$}
\label{fig:ratio_experiments_samples}
\end{figure}

In Figure \ref{fig:ratio_experiments_samples}, we illustrate a random sample from the Cityscapes dataset and its results given different training dataset configurations. Refer to supplementary material for more results.

\subsection{How unsupervised Image-to-Image translation algorithms uses implicit pair information}
It has been shown that when various GAN models optimize their objective function they essentially learn to minimize some $f$-divergence between the distribution of the generated samples and that of the real data \cite{Nowozin2016fGANTG, JolicoeurMartineau2019OnRF}:

\begin{equation}
D_f(P_{data} || P_{fake}) = 
\int_{\chi} P_{data}(x)f \left(\dfrac{P_{fake}(x)}{P_{data}(x)}\right)dx
\label{eq:fdivergence}
\end{equation}

where $P_{data}, P_{fake}$ are the distributions of the discriminator in relation to the real and generated samples respectively, $f$ is a convex function with $f(1)=0$ and $D_f \geq 0$, i.e. a minima of $D_f$ is when $\forall x \; P_{data}(x) = P_{fake}(x)$.

For example the original GAN model \cite{NIPS2014_5423} can be described as minimizing:

\begin{equation}
\int_{\chi} log \dfrac{2P_{data}}{P_{data} + P_{fake}} P_{data} \; + \; 
\;  log \dfrac{2P_{fake}}{P_{data}+P_{fake}} P_{fake}
\end{equation}

Where as can be seen, if for a given $x$ we reduce the term $|P_{fake}(x) - P_{data}(x)|$ we get better minimization of the objective.

Now consider how the samples $a_i \in A, T(a_i) \in B$ affect this minimization process. By enriching the data with $T(a_i)$ allow the discriminator to be more discerning in its estimation of $P_{data}(T(a_i))$. Assuming a passable generator would translate $a_i$ to the neighborhood of $T(a_i)$ by generating $G_A(a_i) \approx T(a_i)$, and using Occam's Razor, we can expect the enriched discriminator to provide better gradients for $G_A(a_i)$ w.r.t. the transformation we wish to learn.
Our experimental results suggest that although the injection of $T(a_i)$ into the dataset does not constrain the algorithm to use it for improving the model, it nevertheless does.

The implicit pairs convey to the discriminator precise information in the region of the translated image $G_A(a_i)$ which is used to direct the learning process. At the same time, using only paired samples constrains the space that is explored by the model and coupled with the fine-tuned information leads to overfitting.


We would like to note two limitations for this approach. First, when using a generator which exhibits pathological behaviors, such as mode collapse, adding such data points would not improve the results since the added data point is either outside of the pathology and is not used with respect to the generator or it is within the pathology and can only reinforce the pathology.
Similarly, if the generator is very poor and translates $a_i$ to some point very far from $T(a_i)$ we can expect the beneficial signal of the paired sample to be diminished in favor of closer samples.

\section{Data augmentation using implicit pseudo-pairs}
\label{sec:method}

\subsection{Implicit pseudo-pairs}
In Section \ref{sec:implicit_pairs} we have shown that unsupervised Image-to-Image translation models can use the information presented implicitly by the occurrence of pairs in the dataset during training. Unfortunately, there are many situations where obtaining pairs, even implicitly, is hard or impossible, and in order to use pairing information without first obtaining pairs we will need to create them. 

With that aim in mind we extend the explanation of how implicit pairs are beneficial when training unsupervised Image-to-Image translation models.
From Equation \ref{eq:fdivergence} we surmised that the learning process will utilize the existence of $T(a_i)$ to enhance translation of $a_i$ into $G(a_i) \approx T(a_i)$. But the majority of translation tasks are not interested in translation between completely disparate domains, instead focused on translation in limited dimensions (hair color, color space, image modality, etc.). In that case we can describe $a_i$ by its two parts: dimensions which are affected by the translation and dimensions which are not.

\begin{figure}
\centering%
	\jsubfig{\includegraphics[height=2.4cm,]{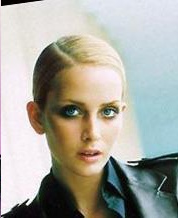}}
	{}%
	\hspace{1pt}
	\jsubfig{\includegraphics[height=2.4cm,]{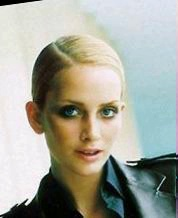}}
	{}%
	\hfill%
	\jsubfig{\includegraphics[height=2.4cm,]{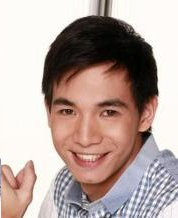}}
	{}%
	\hspace{1pt}
	\jsubfig{\includegraphics[height=2.4cm,]{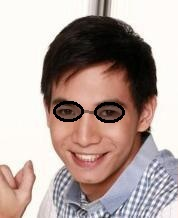}}
	{}%
\\

	\jsubfig{\includegraphics[height=2.4cm,]{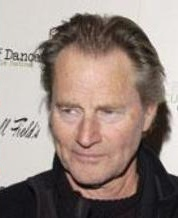}}
	{}%
	\hspace{1pt}
	\jsubfig{\includegraphics[height=2.4cm,]{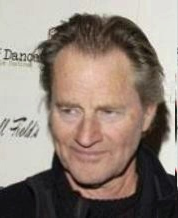}}
	{}%
	\hfill%
	\jsubfig{\includegraphics[height=2.4cm,]{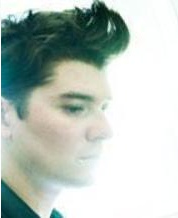}}
	{}%
	\hspace{1pt}
	\jsubfig{\includegraphics[height=2.4cm,]{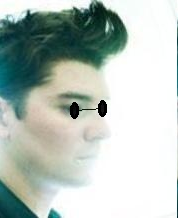}}
	{}%
\\
	\jsubfig{\includegraphics[height=2.4cm,]{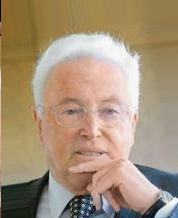}}
	{}%
	\hspace{1pt}
	\jsubfig{\includegraphics[height=2.4cm,]{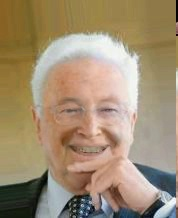}}
	{}%
	\hfill%
	\jsubfig{\includegraphics[height=2.4cm,]{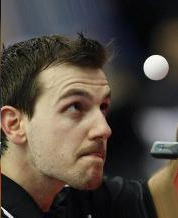}}
	{}%
	\hspace{1pt}
	\jsubfig{\includegraphics[height=2.4cm,]{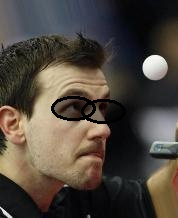}}
	{}%
\\
	{(a)}%
	\hspace{110pt}
	{(b)}%
\\

\caption{{\bfseries illustration of pseudo-pairs.} (a) Pseudo-smiling and neutral faces (b) Pseudo-eyeglasses and faces without eyeglasses.
}

\label{fig:pseudo_pairs_samples}
\end{figure}

Assume we are given a transformation $T'$ s.t. $T'$ only affects the dimensions which are affected by $T$, and that we use $T'$ to create synthetic image samples $(a_i, T'(a_i))$ to create implicit pseudo-pairs and train an unsupervised Image-to-Image translation model on. Consider how the pseudo-pairs affect the training process: the unaffected dimensions (i.e. the parts of the image which are still a "real pairing") would be improved just as in the real implicit pairs case. The affected dimensions would depend on how close $T'$ is to $T$ but as we will show in the following sections, even a poor estimation of $T$ could improve the translation for the changed dimensions.
Therefore we suggest a data augmentation method in which pseudo-pairs are synthesized as in Figure \ref{fig:pseudo_pairs_samples} and used to enrich an existing dataset implicitly.

In reality the cases where we cannot easily obtain an approximating transformation $T'$ are more interesting, therefore we will focus on cases where we can obtain an approximation of the \textit{inverse} transformation, $M \approx T^{-1} : B \rightarrow A$. In other words, we want to improve the learned transformation $G_A : A \rightarrow B$ by introducing synthetic samples $M(b)$ to domain $A$. Given such a generative model $M$ we evaluate how effective these imperfect pseudo-pairs are by extending our experiments to pseudo-paired datasets where the pairing is carried out between \textit{generated} pseudo-samples in domain \textsc{A} and real samples in domain \textsc{B}. See Figure \ref{fig:pseudo_pairs_samples} for an illustration of pseudo-pairs in different datasets.

\begin{figure}
\centering
\includegraphics[width=1.0\columnwidth]{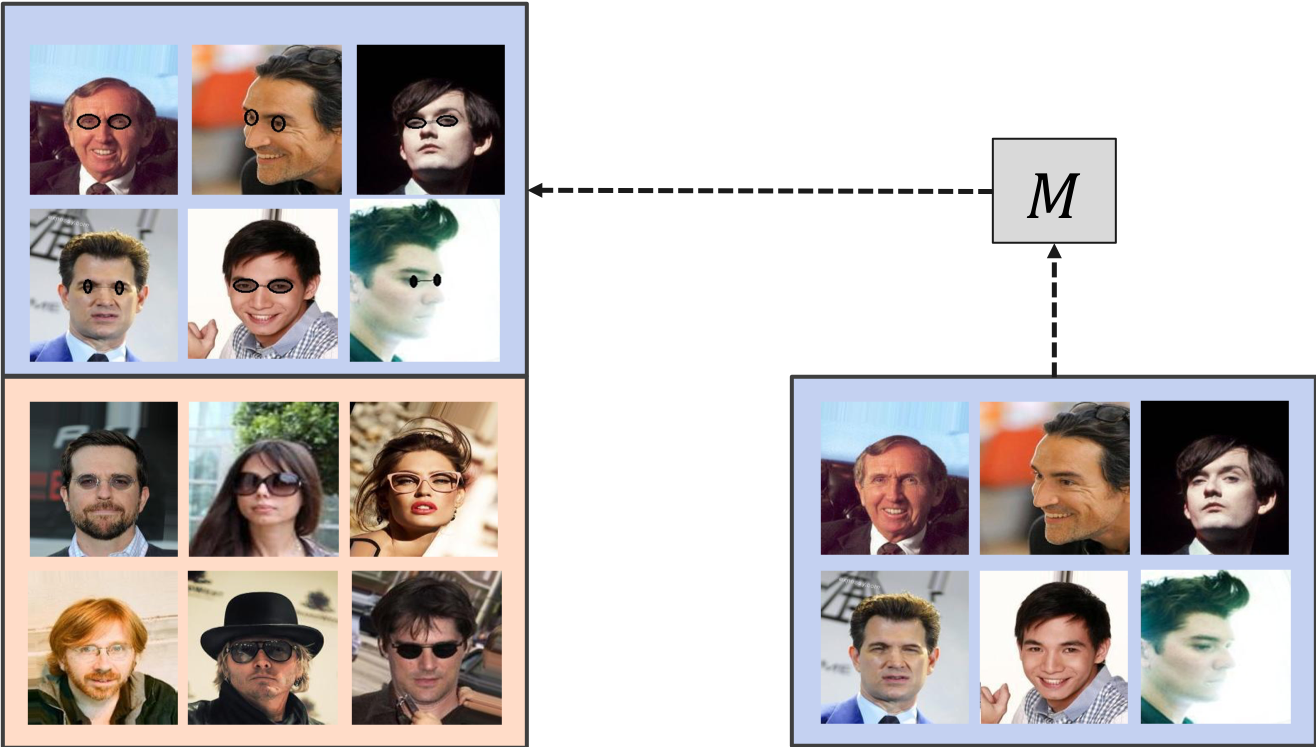}
\\
A \hspace{130pt} B

\caption{Learning using implicit pseudo-pairs with $ \alpha = 0.5$. Given a model $M$, we augment domain $A$ by generating approximations using samples from domain $B$.}
\label{fig:method_overview}
\end{figure}

Figure \ref{fig:method_overview} provides an overview of our approach in this setting. Given an unpaired dataset, we construct an $\alpha$-pseudo-paired dataset using a generative model $M$ to inject pseudo-samples to the unpaired sets. 

In the following section, we report on experiments that show that implicit pseudo-pairs boost the performance in the unpaired setting, and that using them as explicit pairs in a paired setting is significantly less effective. 

\section{Experiments and Results}

\label{sec:pseudo_pairs_subsection}

In the following section we demonstrate through experiments the efficacy of implicit pseudo-pairs when training unsupervised Image-to-Image translation algorithms.

\subsection{Sample synthesis}
We use model-based generative techniques to estimate the inverse transformation to the one we are trying to learn, $M \approx T^{-1} : B \rightarrow A$. For each sample in domain $B$ we generate a paired pseudo-sample to augment the samples in domain $A$. This creates pseudo-paired datasets with a 50\% pairing ratio. An overview of this method is shown in Figure \ref{fig:method_overview}.

\paragraph{Dataset.}
For the experiments on pseudo-pairs, we use the CelebA dataset \cite{liu2015faceattributes}. We generate two different types of pseudo $\alpha$-paired datasets on which we evaluate our method: (i) faces with and without eyeglasses and (ii) smiling and neutral faces. The datasets can be obtained using the labeling information available for each image in the CelebA dataset. In (i), we generate pseudo-samples with eyeglasses using simple heuristics. Using the available facial landmarks, we generate ellipses around the eyes by sampling a random height $h$ in the range of [10, 25] pixels, a random width in the range of [$h / 2$ , $2 \cdot h$] and a transparency coefficient in the range [0.1, 1.0]. The two ellipses are connected by a line with the same transparency and with width in the range of [$h / 5$ , $h / 2$].
In (ii), we use the technique of Averbuch et al. \cite{averbuch2017bringing} to generate smiling pseudo-samples. It is important to note that in both cases, it is significantly more challenging to generate clean samples in the inverse direction. See Figure \ref{fig:pseudo_pairs_samples} for an illustration of pseudo-pairs in both types of datasets. 

\paragraph{Evaluation metrics.}
As described before, Image-to-Image translation is often a translation in some of the image dimensions but not in all of them. This is clearly observed in tasks related to the CelebA dataset where one or more attributes of the image are translated (hair color, existence of eyeglasses, gender, etc.) while the identity of the person is expected to remain the same after the translation. This leads us to frame the evaluation of our results in terms of task completion (how well the 
translation of the attributes in question was done) and identity preservation (how well other attributes were conserved).

According to our description of implicit pairs we would expect that using pseudo-pairs would improve the conservation of the identity after the transformation as it is part of the image dimensions which should not be affected by the model-based techniques outlined above.

Following previous works that use the MSE in representation space as either a perceptual or an identity loss term \cite{DBLP:journals/corr/abs-1711-10352, DBLP:journals/corr/AntipovBD17, DBLP:journals/corr/LedigTHCATTWS16, DBLP:journals/corr/JohnsonAL16, DBLP:journals/corr/WangXWT17}, we will use a representation-space similarity measure which we denote \texttt{InfoSIM}, to measure the preservation of information not related to the task between the input sample and its generated counterpart. In the case of measuring how well the facial identity is preserved after the translation we use the representations learned by the OpenFace network \cite{amos2016openface}. This network is trained for facial recognition and is invariant to transient features, such as smiling or wearing eyeglasses. To measure the similarity we use the MSE between the representation of the input and output images.  

To evaluate task completion we perform a user study where human participants evaluated the task completion of several variants of our method.

\paragraph{Experimental setup.}
In our experiments, we sample $1000$ unpaired samples from CelebA \cite{liu2015faceattributes} from each domain, which we augment with another $1000$ samples according to the augmentation method used in the specific experiment. The resolution of the images is 128X128. Unless stated otherwise, all of the experiments were done using the CycleGAN model described above.

\paragraph{Comparing implicit pseudo-pairs to baseline methods.} \label{basic_pseudo_pairs_experiments}
We evaluate our pseudo-pairs augmentation technique against three augmentation baselines: (i) no-augmentation, (ii) pseudo-unpaired augmentation and (iii) \textit{natural} augmentation of real images belonging to the corresponding domain. In (i), we do not augment the basic dataset configuration with any samples. In (ii), we augment the basic dataset configuration with pseudo-samples whose paired real samples are not in the dataset. In (iii), we simply augment the basic dataset with more real images, sampled from the full dataset.

\begin
{figure}
\centering %

 \jsubfig{\includegraphics[height=1.55cm]{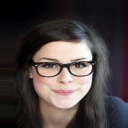}}
 	{}%
 	\hspace{3pt}
 	\jsubfig{\includegraphics[height=1.55cm]{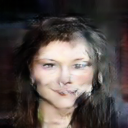}}
 	{}%
 	\hfill%
 	\jsubfig{\includegraphics[height=1.55cm]{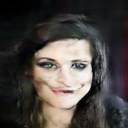}}
 	{}%
  	\hfill%
 	\jsubfig{\includegraphics[height=1.55cm]{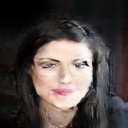}}
 	{}%
  	\hfill%
  	\hspace{3pt}
 	\jsubfig{\includegraphics[height=1.55cm]{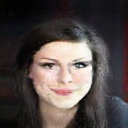}}
 	{}%
 	
 	 \jsubfig{\includegraphics[height=1.55cm]{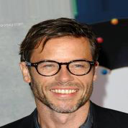}}
 	{}%
 	\hspace{3pt}
 	\jsubfig{\includegraphics[height=1.55cm]{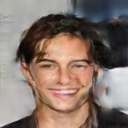}}
 	{}%
 	\hfill%
 	\jsubfig{\includegraphics[height=1.55cm]{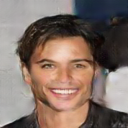}}
 	{}%
  	\hfill%
 	\jsubfig{\includegraphics[height=1.55cm]{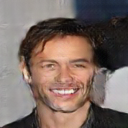}}
 	{}%
  	\hfill%
  	\hspace{3pt}
 	\jsubfig{\includegraphics[height=1.55cm]{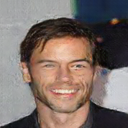}}
 	{}%


\jsubfig{\includegraphics[height=1.55cm]{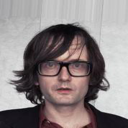}}
	{}%
	\hspace{3pt}
	\jsubfig{\includegraphics[height=1.55cm]{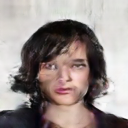}}
	{}%
	\hfill%
	\jsubfig{\includegraphics[height=1.55cm]{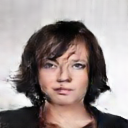}}
	{}%
 	\hfill%
	\jsubfig{\includegraphics[height=1.55cm]{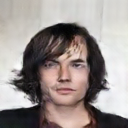}}
	{}%
 	\hfill%
 	\hspace{3pt}
	\jsubfig{\includegraphics[height=1.55cm]{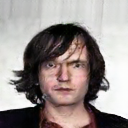}}
	{}%

\jsubfig{\includegraphics[height=1.55cm]{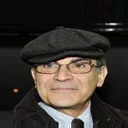}}
	{}%
	\hspace{3pt}
	\jsubfig{\includegraphics[height=1.55cm]{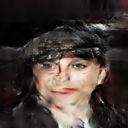}}
	{}%
	\hfill%
	\jsubfig{\includegraphics[height=1.55cm]{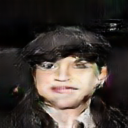}}
	{}%
 	\hfill%
	\jsubfig{\includegraphics[height=1.55cm]{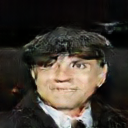}}
	{}%
 	\hfill%
 	\hspace{3pt}
	\jsubfig{\includegraphics[height=1.55cm]{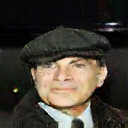}}
	{}%

\jsubfig{\includegraphics[height=1.55cm]{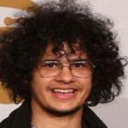}}
	{}%
	\hspace{3pt}
	\jsubfig{\includegraphics[height=1.55cm]{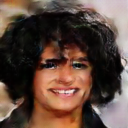}}
	{}%
	\hfill%
	\jsubfig{\includegraphics[height=1.55cm]{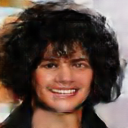}}
	{}%
 	\hfill%
	\jsubfig{\includegraphics[height=1.55cm]{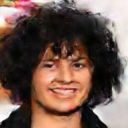}}
	{}%
 	\hfill%
 	\hspace{3pt}
	\jsubfig{\includegraphics[height=1.55cm]{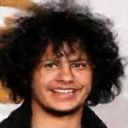}}
	{}%


\jsubfig{\includegraphics[height=1.55cm]{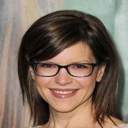}}
	{\small{Source}}%
	\hspace{3pt}
	\jsubfig{\includegraphics[height=1.55cm]{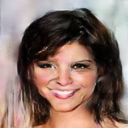}}
	{\small{baseline}}%
	\hfill%
	\jsubfig{\includegraphics[height=1.55cm]{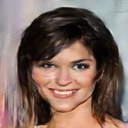}}
	{\small{+natural}}%
 	\hfill%
	\jsubfig{\includegraphics[height=1.55cm]{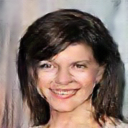}}
	{\small{+unpaired}}%
 	\hfill%
 	\hspace{3pt}
	\jsubfig{\includegraphics[height=1.55cm]{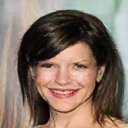}}
	{\small{+paired}}%
\\
\vspace{3pt}
\caption{\textit{Eyeglass removal} results using different dataset configurations. Above we illustrate our results (on the right) compared against three augmentation baselines, described in Section \ref{sec:pseudo_pairs_subsection}}

\label{fig:quality_eyeglasses}
\end{figure}

\begin
{figure}
\centering %

\jsubfig{\includegraphics[height=1.55cm]{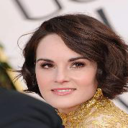}}
	{}%
	\hspace{3pt}
	\jsubfig{\includegraphics[height=1.55cm]{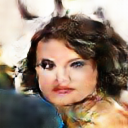}}
	{}%
	\hfill%
	\jsubfig{\includegraphics[height=1.55cm]{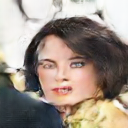}}
	{}%
 	\hfill%
	\jsubfig{\includegraphics[height=1.55cm]{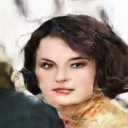}}
	{}%
 	\hfill%
	\hspace{3pt}
	\jsubfig{\includegraphics[height=1.55cm]{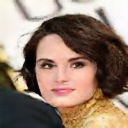}}
	{}%

\jsubfig{\includegraphics[height=1.55cm]{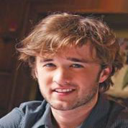}}
	{}%
	\hspace{3pt}
	\jsubfig{\includegraphics[height=1.55cm]{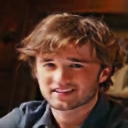}}
	{}%
	\hfill%
	\jsubfig{\includegraphics[height=1.55cm]{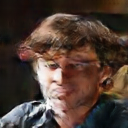}}
	{}%
 	\hfill%
	\jsubfig{\includegraphics[height=1.55cm]{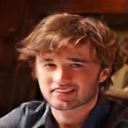}}
	{}%
 	\hfill%
	\hspace{3pt}
	\jsubfig{\includegraphics[height=1.55cm]{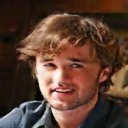}}
	{}%
	
\jsubfig{\includegraphics[height=1.55cm]{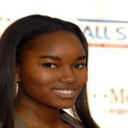}}
	{}%
	\hspace{3pt}
	\jsubfig{\includegraphics[height=1.55cm]{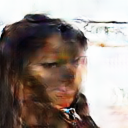}}
	{}%
	\hfill%
	\jsubfig{\includegraphics[height=1.55cm]{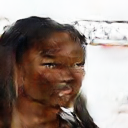}}
	{}%
 	\hfill%
	\jsubfig{\includegraphics[height=1.55cm]{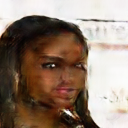}}
	{}%
 	\hfill%
	\hspace{3pt}
	\jsubfig{\includegraphics[height=1.55cm]{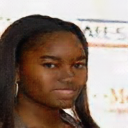}}
	{}%


 \jsubfig{\includegraphics[height=1.55cm]{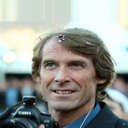}}
 	{}%
 	\hspace{3pt}
 	\jsubfig{\includegraphics[height=1.55cm]{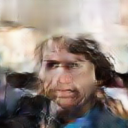}}
 	{}%
 	\hfill%
 	\jsubfig{\includegraphics[height=1.55cm]{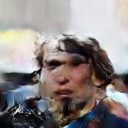}}
 	{}%
  	\hfill%
 	\jsubfig{\includegraphics[height=1.55cm]{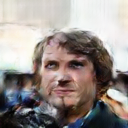}}
 	{}%
  	\hfill%
 	\hspace{3pt}
 	\jsubfig{\includegraphics[height=1.55cm]{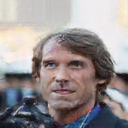}}
 	{}%


\jsubfig{\includegraphics[height=1.55cm]{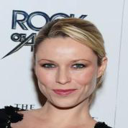}}
	{}%
	\hspace{3pt}
	\jsubfig{\includegraphics[height=1.55cm]{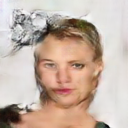}}
	{}%
	\hfill%
	\jsubfig{\includegraphics[height=1.55cm]{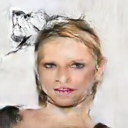}}
	{}%
 	\hfill%
	\jsubfig{\includegraphics[height=1.55cm]{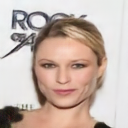}}
	{}%
 	\hfill%
	\hspace{3pt}
	\jsubfig{\includegraphics[height=1.55cm]{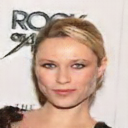}}
	{}%

\jsubfig{\includegraphics[height=1.55cm]{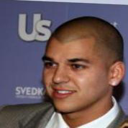}}
	{\small{Source}}%
	\hspace{3pt}
	\jsubfig{\includegraphics[height=1.55cm]{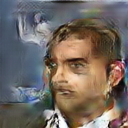}}
	{\small{baseline}}%
	\hfill%
	\jsubfig{\includegraphics[height=1.55cm]{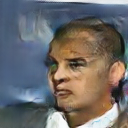}}
	{\small{+natural}}%
 	\hfill%
	\jsubfig{\includegraphics[height=1.55cm]{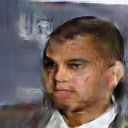}}
	{\small{+unpaired}}%
 	\hfill%
	\hspace{3pt}
	\jsubfig{\includegraphics[height=1.55cm]{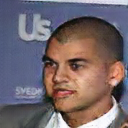}}
	{\small{+paired}}%

\vspace{3pt}
\caption{Results for the  \textit{smile removal} task using different dataset configurations. Above we illustrate our results (on the right) compared against three augmentation baselines, described in Section \ref{sec:pseudo_pairs_subsection}.
 }

\label{fig:quality_smiling}
\end{figure}

The identity preservation results for the baseline methods are reported in Table \ref{table:basic_experiments_preservation}. Figures  \ref{fig:quality_eyeglasses}, \ref{fig:quality_smiling} demonstrate the qualitative results of these experiments. As the results illustrate, using implicit pseudo-pairs improves the quality of the translation while better preserving the facial identity. For instance, it is especially noticeable that using implicit pseudo-pairs introduces fewer artifacts in comparison to the other approaches.

\begin{table}[h!]
\centering
\begin{tabular}{ p{1.2cm}|p{1.0cm}p{1.0cm}p{1.0cm}p{1.0cm} } 
 Task & 
 \multicolumn{1}{c}{\textbf{ours}} & \multicolumn{1}{c}{\textbf{(i)}} & \multicolumn{1}{c}{\textbf{(ii)}}
 & \multicolumn{1}{c}{\textbf{(iii)}}
 \\
\hline
 Smile & \textbf{0.00160}   & 0.00365 &  0.00282 &   0.00372 \\
 Eyeglass & \textbf{0.00181}    & 0.00482 & 0.00313 &   0.00418\\
 
\end{tabular}
\caption{\texttt{InfoSIM} comparison between the baseline augmentation methods described in Subsection \ref{basic_pseudo_pairs_experiments}. Lower is better.}
\label{table:basic_experiments_preservation}
\end{table}

\paragraph{Pseudo-pairs ratio analysis.} 
In Section \ref{sec:implicit_pairs}, we demonstrated that having different ratios of pairs in the dataset can have a significant effect on the results. Here we continue this line of inquiry by evaluating the effect different ratios of pseudo-pairs have. We test the following $\alpha$-paired configurations: $\alpha = 0.25,0.5,0.75,1.0$. To create a \textbf{$\alpha=0.25$} pseudo-paired dataset, half of the augmentation samples are paired and the other half are unpaired. To create datasets with a pairing ratio higher than 50\% we remove domain $A$ samples from the initial dataset and augment with more pseudo-pairs. For example, the \textbf{0.75}-pseudo-paired dataset has 500 unpaired real samples augmented with 1500 pseudo-pairs. 

\begin
{figure}
\centering %


\jsubfig{\includegraphics[height=1.55cm]{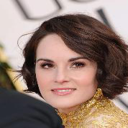}}
	{}%
	\hspace{3pt}
	\jsubfig{\includegraphics[height=1.55cm]{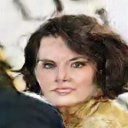}}
	{}%
	\hfill%
	\jsubfig{\includegraphics[height=1.55cm]{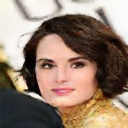}}
	{}%
 	\hfill%
	\jsubfig{\includegraphics[height=1.55cm]{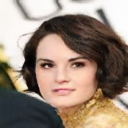}}
	{}%
 	\hfill%
	\jsubfig{\includegraphics[height=1.55cm]{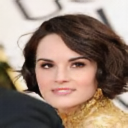}}
	{}%

\jsubfig{\includegraphics[height=1.55cm]{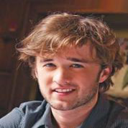}}
	{}%
	\hspace{3pt}
	\jsubfig{\includegraphics[height=1.55cm]{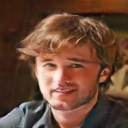}}
	{}%
	\hfill%
	\jsubfig{\includegraphics[height=1.55cm]{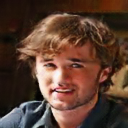}}
	{}%
 	\hfill%
	\jsubfig{\includegraphics[height=1.55cm]{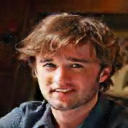}}
	{}%
 	\hfill%
	\jsubfig{\includegraphics[height=1.55cm]{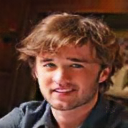}}
	{}%

\jsubfig{\includegraphics[height=1.55cm]{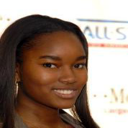}}
	{}%
	\hspace{3pt}
	\jsubfig{\includegraphics[height=1.55cm]{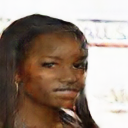}}
	{}%
	\hfill%
	\jsubfig{\includegraphics[height=1.55cm]{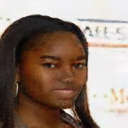}}
	{}%
 	\hfill%
	\jsubfig{\includegraphics[height=1.55cm]{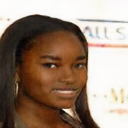}}
	{}%
 	\hfill%
	\jsubfig{\includegraphics[height=1.55cm]{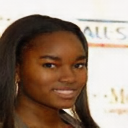}}
	{}%

 	{}%

 \jsubfig{\includegraphics[height=1.55cm]{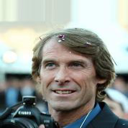}}
 	{}%
 	\hspace{3pt}
 	\jsubfig{\includegraphics[height=1.55cm]{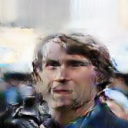}}
 	{}%
 	\hfill%
 	\jsubfig{\includegraphics[height=1.55cm]{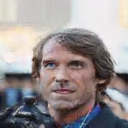}}
 	{}%
  	\hfill%
 	\jsubfig{\includegraphics[height=1.55cm]{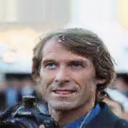}}
 	{}%
  	\hfill%
 	\jsubfig{\includegraphics[height=1.55cm]{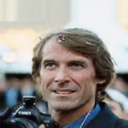}}
 	{}%

\jsubfig{\includegraphics[height=1.55cm]{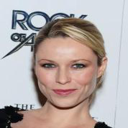}}
	{}%
	\hspace{3pt}
	\jsubfig{\includegraphics[height=1.55cm]{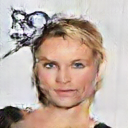}}
	{}%
	\hfill%
	\jsubfig{\includegraphics[height=1.55cm]{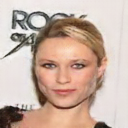}}
	{}%
 	\hfill%
	\jsubfig{\includegraphics[height=1.55cm]{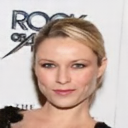}}
	{}%
 	\hfill%
	\jsubfig{\includegraphics[height=1.55cm]{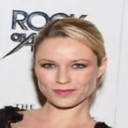}}
	{}%

\jsubfig{\includegraphics[height=1.55cm]{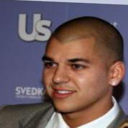}}
	{\small{Source}}%
	\hspace{3pt}
	\jsubfig{\includegraphics[height=1.55cm]{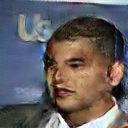}}
	{\small{25\%}}%
	\hfill%
	\jsubfig{\includegraphics[height=1.55cm]{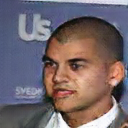}}
	{\small{50\%}}%
 	\hfill%
	\jsubfig{\includegraphics[height=1.55cm]{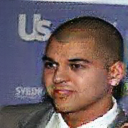}}
	{\small{75\%}}%
 	\hfill%
	\jsubfig{\includegraphics[height=1.55cm]{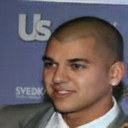}}
	{\small{100\%}}%

\vspace{3pt}
\caption{Pseudo-pair ratio analysis for the  \textit{smile removal} task. Above we illustrate a few results using various pairing ratios. As the figure illustrates, using a 50\% pairing configuration yields identity-preserving results which still perform the task (smile removal in this case) better than a higher pairing ratio.  }

\label{fig:smiling_pairing_ratios}
\end{figure}

\begin
{figure}

\centering %

\jsubfig{\includegraphics[height=1.55cm]{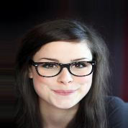}}
{}%
\hspace{3pt}
\jsubfig{\includegraphics[height=1.55cm]{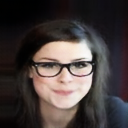}}
{}%
\hfill%
\jsubfig{\includegraphics[height=1.55cm]{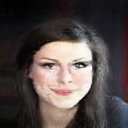}}
{}%
\hfill%
\jsubfig{\includegraphics[height=1.55cm]{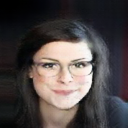}}
{}%
\hfill%
\jsubfig{\includegraphics[height=1.55cm]{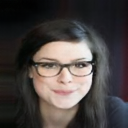}}
{}%

\jsubfig{\includegraphics[height=1.55cm]{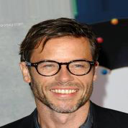}}
{}%
\hspace{3pt}
\jsubfig{\includegraphics[height=1.55cm]{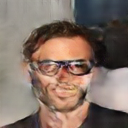}}
{}%
\hfill%
\jsubfig{\includegraphics[height=1.55cm]{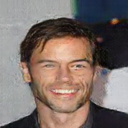}}
{}%
\hfill%
\jsubfig{\includegraphics[height=1.55cm]{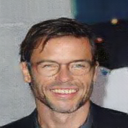}}
{}%
\hfill%
\jsubfig{\includegraphics[height=1.55cm]{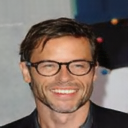}}
{}%

\jsubfig{\includegraphics[height=1.55cm]{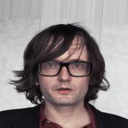}}
	{}%
	\hspace{3pt}
	\jsubfig{\includegraphics[height=1.55cm]{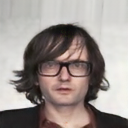}}
	{}%
	\hfill%
	\jsubfig{\includegraphics[height=1.55cm]{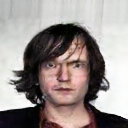}}
	{}%
 	\hfill%
	\jsubfig{\includegraphics[height=1.55cm]{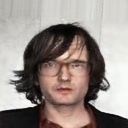}}
	{}%
 	\hfill%
	\jsubfig{\includegraphics[height=1.55cm]{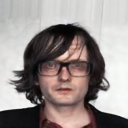}}
	{}%

\jsubfig{\includegraphics[height=1.55cm]{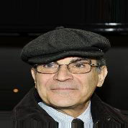}}
	{}%
	\hspace{3pt}
	\jsubfig{\includegraphics[height=1.55cm]{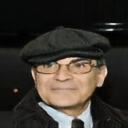}}
	{}%
	\hfill%
	\jsubfig{\includegraphics[height=1.55cm]{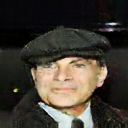}}
	{}%
 	\hfill%
	\jsubfig{\includegraphics[height=1.55cm]{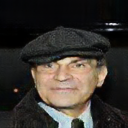}}
	{}%
 	\hfill%
	\jsubfig{\includegraphics[height=1.55cm]{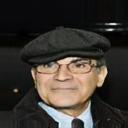}}
	{}%

\jsubfig{\includegraphics[height=1.55cm]{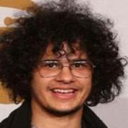}}
	{}%
	\hspace{3pt}
	\jsubfig{\includegraphics[height=1.55cm]{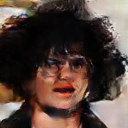}}
	{}%
	\hfill%
	\jsubfig{\includegraphics[height=1.55cm]{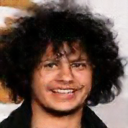}}
	{}%
 	\hfill%
	\jsubfig{\includegraphics[height=1.55cm]{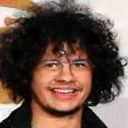}}
	{}%
 	\hfill%
	\jsubfig{\includegraphics[height=1.55cm]{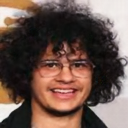}}
	{}%

\jsubfig{\includegraphics[height=1.55cm]{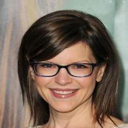}}
	{\small{Source}}%
	\hspace{3pt}
	\jsubfig{\includegraphics[height=1.55cm]{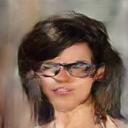}}
	{\small{25\%}}%
	\hfill%
	\jsubfig{\includegraphics[height=1.55cm]{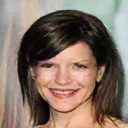}}
	{\small{50\%}}%
 	\hfill%
	\jsubfig{\includegraphics[height=1.55cm]{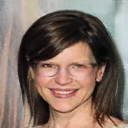}}
	{\small{75\%}}%
 	\hfill%
	\jsubfig{\includegraphics[height=1.55cm]{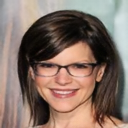}}
	{\small{100\%}}%

\vspace{3pt}
\caption{Pseudo-pair ratio analysis for the  \textit{eyeglass removal} task. Above we illustrate a few randomly selected results using various pairing ratios. As the figure illustrates, using a 50\% pairing configuration yields reasonable identity-preserving results while perform the task (eyeglass removal in this case) better than a higher pairing ratio.  }

\label{fig:eyeglasses_pairing_ratios}
\end{figure}

The identity preservation results for these experiments are reported in Table \ref{table:pairing_ratios_preservation}. The results clearly demonstrate that the more pairs we have in the dataset, the better the identity is preserved. The qualitative results for these experiments are demonstrated in Figures \ref{fig:smiling_pairing_ratios}, \ref{fig:eyeglasses_pairing_ratios}. 
As the figures illustrate, having more pairs allows us to better preserve the facial identity which supports our hypothesis that implicit pairs enhance the preservation of dimensions not directly related to the transformation used. At the same time, it is also apparent that higher pairing ratio leads to poor task completion as the model is exposed to more pseudo-pairs and fewer examples of the real domain, and is thus less able to generalize to the real task. This is especially pronounced in the \textit{smile removal} task as the generation model $M$ is based on a closed set of smile templates and generalizing from such a limited set is hard.

\paragraph{Task completion user study.}
To further quantify the success of different ratios, we conduct a user study in which participants are presented with translated results generated using models trained with $0\%, 50\%$ and $100\%$ pseudo-paired datasets and asked which model completes the task better. To allow for fine-grained comparison, the participants are shown the results of only two models at a time (or one model and the source image) to choose one, both or none if both are equally good or bad.

We had a total of 43 participants which completed separate studies for 50 eyeglass wearing and 41 smiling samples. 

In Table \ref{table:userstudy} we report the rate by which participants preferred one model over the other. It is clear that using a $50\%$ pseudo-paired yields the best results in terms of task completion. 

\begingroup
\setlength{\tabcolsep}{4.0pt}
\begin{table}[h!]
\centering
\begin{tabular}{ p{1.2cm}|p{0.2cm}p{0.2cm}p{0.2cm}p{0.2cm} } 
 Task & 
 \multicolumn{1}{c}{0.25-Paired} & \multicolumn{1}{c}{0.5-Paired} & \multicolumn{1}{c}{0.75-Paired}
 & \multicolumn{1}{c}{1-Paired}
 \\
\hline
 Smile &  0.00293 & 0.00160 & 0.00093 & \textbf{0.00025} \\
 Eyeglass &  0.00153 & 0.00181 & 0.00110 & \textbf{0.00025} \\
 
\end{tabular}
\caption{\texttt{InfoSIM} values for pairing ratios experiments. Lower is better.}
\label{table:pairing_ratios_preservation}
\end{table}
\endgroup
 

\vspace{-1em}
\begin{table}[h!]
\centering
\begin{tabular}{ p{3.0cm}|p{1.0cm}p{1.0cm}p{1.0cm} } 

 \backslashbox[35mm]{Preferred}{Rejected} & 
 \multicolumn{1}{c}{0\%} & \multicolumn{1}{c}{50\%} & \multicolumn{1}{c}{100\%}
 
 \\
\hline
 0\% (Smile) & -   & 0.257 &  0.486 \\
 50\% (Smile) & \textbf{0.742}   & - &  \textbf{0.663} \\
 100\% (Smile) & 0.513   & 0.336 &  - \\

\hline
 0\% (Eyeglass) & -   & 0.264 &  0.915 \\
 50\% (Eyeglass) & \textbf{0.735}   & - & \textbf{1.0} \\
 100\% (Eyeglass) & 0.084   & 0.0 &  - \\
 
 \hline
\end{tabular}
\caption{Task completion preference rate according to user study}
\label{table:userstudy}
\end{table}


\paragraph{Pseudo-pairs in different image-to-image translation settings.}
\begin
{figure}
\centering %

\jsubfig{\includegraphics[height=2.7cm]{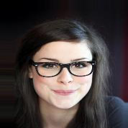}}
{}%
\hspace{3pt}
\jsubfig{\includegraphics[height=2.7cm]{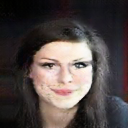}}
{}%
\hfill%
\jsubfig{\includegraphics[height=2.7cm]{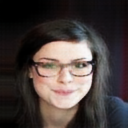}}
{}%

\jsubfig{\includegraphics[height=2.7cm]{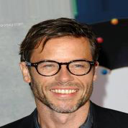}}
{}%
\hspace{3pt}
\jsubfig{\includegraphics[height=2.7cm]{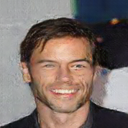}}
{}%
\hfill%
\jsubfig{\includegraphics[height=2.7cm]{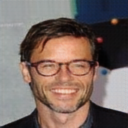}}
{}%

\jsubfig{\includegraphics[height=2.7cm]{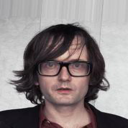}}
	{}%
	\hspace{3pt}
	\jsubfig{\includegraphics[height=2.7cm]{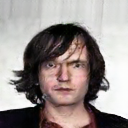}}
	{}%
	\hfill%
	\jsubfig{\includegraphics[height=2.7cm]{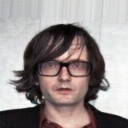}}
	{}%

\jsubfig{\includegraphics[height=2.7cm]{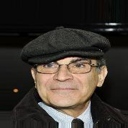}}
	{}%
	\hspace{3pt}
	\jsubfig{\includegraphics[height=2.7cm]{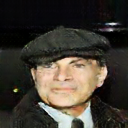}}
	{}%
	\hfill%
	\jsubfig{\includegraphics[height=2.7cm]{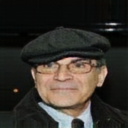}}
	{}%

\jsubfig{\includegraphics[height=2.7cm]{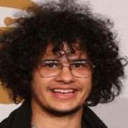}}
	{}%
	\hspace{3pt}
	\jsubfig{\includegraphics[height=2.7cm]{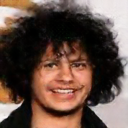}}
	{}%
	\hfill%
	\jsubfig{\includegraphics[height=2.7cm]{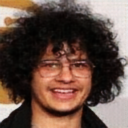}}
	{}%

\jsubfig{\includegraphics[height=2.7cm]{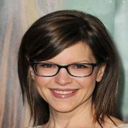}}
	{\small{Source}}%
	\hspace{3pt}
	\jsubfig{\includegraphics[height=2.7cm]{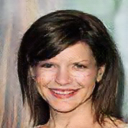}}
	{\small{Implicit}}%
	\hfill%
	\jsubfig{\includegraphics[height=2.7cm]{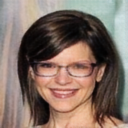}}
	{\small{Explicit}}%
\\
\vspace{3pt}
\caption{\textit{Eyeglass removal} with pseudo-pairs in different settings. Above we illustrate a few randomly selected results by models trained using either an explicit \cite{DBLP:journals/corr/IsolaZZE16} and implicit  \cite{DBLP:journals/corr/ZhuPIE17} settings.
As the figure illustrates, an  implicit setting leads to the best and most consistent results.}

\label{fig:eyeglasses_training_regimes}
\end{figure}

In previous experiments we have used the generated pseudo-pairs in an implicit fashion. To understand more fully the effect the pairs have on training of models we further experiment with using them in an explicit setting. For explicit training we use the previously mentioned Pix2Pix model \cite{DBLP:journals/corr/IsolaZZE16} with the completely pseudo-paired dataset and compare it against the our 0.5-pseudo-paired implicitly trained results.

From the \texttt{InfoSIM} values in Table \ref{table:training_regimes_identity_preservation} and the qualitative results in Figure 8 it is clear that the explicit algorithm barely changes the input, thus achieving a very good identity preservation while not actually completing the task.

We suggest that this happens because the explicit experiment is completely pseudo-paired, i.e. there are no real samples of eyeglass images which leads the model to overfit to the dataset and particularly to the features of the pseudo-glasses which are different from real eyeglasses. This prevents it from generalizing to the real eyeglasses in the test set.
 
This result suggests that as long as the generation model $M$ is not perfect, it would introduce features that the explicit method will overfit on, and the only efficient way to use the pseudo-samples might be in an implicit manner.

\begin{table}
\centering
\begin{tabular}{ p{1.3cm}p{1.3cm} }
 \multicolumn{1}{c}{Implicit} & \multicolumn{1}{c}{Explicit} \\
\hline
 0.00181    & \textbf{0.000342} \\
\end{tabular}
\caption{\texttt{InfoSIM} values for explicit and implicit experiments on the  \textit{Eyeglass removal} task. Lower is better.}
\label{table:training_regimes_identity_preservation}
\end{table}

\begin{figure}
		\jsubfig{\includegraphics[height=3.1cm]{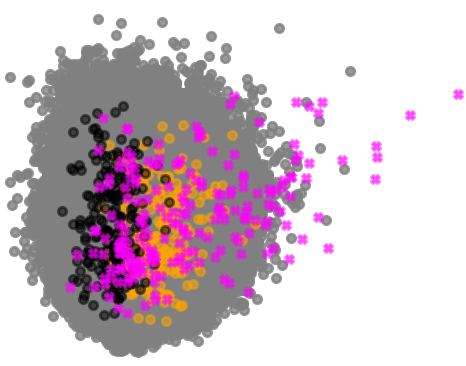}}
		{Space of training data}%
		\hspace{5pt}
		\jsubfig{\includegraphics[height=3.1cm]{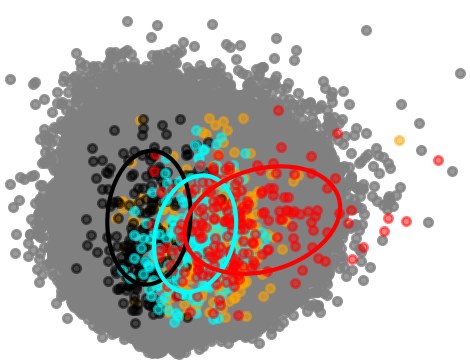}}
		{Space of test data}%

\vspace{3pt}
\caption{Pseudo-pairs visualized on learned expression-related features for the  \textit{smile removal} task. We use PCA to visualize training data (on the left) and test data (on the right). The PCA model is fitted on the entire CelebA dataset (Shown in \gray{gray}). Real smiling and neutral samples are denoted in \orange{orange} and \textbf{black} respectively. Generated augmentation samples are in \fuchsia{fuchsia} and the result samples of translation using the Pix2Pix and CycleGAN algorithms are in \red{red} and \cyan{cyan} respectively with corresponding fitted ellipses. Both models were trained on a 100\% pseudo-paired dataset.}
\label{fig:distribution_analysis}
\end{figure}

\paragraph{Distribution analysis.}

In Section \ref{sec:method} we discussed our hypothesis that by injecting pseudo-pairs into the dataset we add two signals to the learned transformation - The first signal comprises of all of the information in the image that wasn't affected by our synthesis model $M$ and the second by all of the information that was. By looking at the enriched dataset from a perspective of image features we might be able to visualize this effect.

To do so we look at the feature spaces for expressions in celebA. we trained a simple CNN to distinguish between smiling and natural faces on a hold-out set from the dataset, extracted the model representations for the entire celebA dataset, fit a PCA model to them and projected them to 2D space.
We trained Pix2Pix and CycleGAN models on 100\% pseudo-paired dataset and extracted the representation of their results for the test set.

As we can be seen in Figure 9 there is a prominent divergence between the expression representation of the real data (in gray, black and orange) and the representation of the generated samples (in fuchsia). When training on a dataset which comprises solely of generated pseudo-pairs we can see that the divergence is repeated in the results of the Pix2Pix model (in red) but not in the results for the same samples by the unsupervised CycleGAN model (in cyan). We fit ellipses to the results for emphasize.

The similarity in the behavior of the generated data and the results of the Pix2Pix model compared to those of the CycleGAN model suggests that the more powerful and supervised algorithm is able to pick on a signal which the unsupervised model does not, namely that of the fake smiles. This shows that the information the pseudo-samples introduce can be thought of as two separate signals as well as why using explicit pairs can so thoroughly overfit to the data compared to using implicit ones.

\paragraph{Conclusions.}
It is well established that many of the most fundamental human abilities are learned implicitly.
In this work, we analyzed the positive effect of learning with implicitly paired samples in an image-to-image translation problem. We have shown, through numerous experiments and examples, that learning from implicit pairs can effectively guide the network to learn a better mapping, more than additional unpaired or random samples. 

We further analyzed the power of implicit learning using pseudo-pairs. These pseudo-pairs can be obtained automatically either using simple geometric models, as we have shown in the case of faces augmented with eyeglasses, or by more complicated models, such as neutral faces augmented with smiles. In both cases, implicitly providing the network with these pairs yields plausible mappings that better preserve non-task related information.
Additionally, we have shown that datasets augmented with pseudo-pairs can be significantly more effective in an implicit setting than in an explicit one.

The fact that the contribution of the implicit pairs is effective despite their signal being hidden across the dataset, raises the question of what other types of implicit signals a deep neural network may exploit effectively. 
In the future, we believe that exploring the mechanisms by which neural networks learn from implicit signals may shed light on the  understanding of how neural networks learn in general and allow for finer control in the configuration of datasets. 

\clearpage

\bibliographystyle{cas-model2-names}

\bibliography{cas-refs}

\clearpage

\appendix
\section{Implementation details}
\section*{Architectures}
\subsection*{CycleGAN}
In all of our experiments with CycleGAN we have used the vanilla architecture that they have used with 9 residual blocks. Following the naming conventions used in \cite{DBLP:journals/corr/ZhuPIE17}
we express the generator layer parameters as follows:
Define a $7\times 7$ Convolution-InstanceNorm-ReLU layer with $k$ filters and stride 1 as c7s1-$k$, a $3 \times 3$ Convolution-InstnceNorm-ReLU layer with $k$ filters and stride 2 as $dk$, a residual block with $3 \times 3$ convolutional layers with equal numbers of filters on both layers as $Rk$ and a $3 \times 3$ fractional-stridded-Convolution-InstanceNorm-ReLU layer with $k$ filers and stride $1/2$ as $uk$.
For the discriminator we denote a $4 \times 4$ Convolution-InsanceNorm-LeakyReLU layer with $k$ filters and stride 2 with $Ck$.

Using these definitions the generator network can be expressed as: \\
$c7s1-64, d128, d256, R256, R256, R256, \\
R256, R256, R256, R256, R256, R256, \\ u128, 
u64, c7s1-3$

The discriminator network can be similarly expressed as: \\
$C64-C128-C256-C512$.
\\

\subsection*{Pix2Pix}
All experiments involving the Pix2Pix architecture were done using the vanilla version as well. Using the conventions used in \cite{DBLP:journals/corr/IsolaZZE16} for the Pix2Pix network we denote the Convolution-BatchNorm-ReLU layer with $k$ filters as $Ck$ and the Convolution-BatchNorm-Dropout-ReLU layer with 50\% dropout rate as $k$ filters as $CDk$.

The generator is comprised of an encoder expresses as: \\
$C64-C128-C256-C512-C512-C512-C512-C512$ \\
and a decoder expressed as: \\
$CD512-CD512-CD512-C512-C256-C128-C64$

The discriminator can be expressed as: \\
$C64-C128-C256-C512$ \\

\section*{FCN-score}
\paragraph{photo $\rightarrow$ labels}
To convert the generated image to a label matrix we mapped every pixel's rgb value to the label with the lowest mean distance according to the label $\leftrightarrow$ rgb value conversion table provided with the Cityscapes \cite{Cordts2016Cityscapes} dataset.

\paragraph{labels $\rightarrow$ photo}
In this direction we use the same FCN-8 network that was used in \cite{DBLP:journals/corr/IsolaZZE16} to segment the generated image into a label matrix.

\paragraph{} Finally, in either direction we used the evaluation script provided in Zhu's github repository \footnote{https://github.com/junyanz/pytorch-CycleGAN-and-pix2pix}

\section*{CVC-14 dataset}
The dataset contains paired sequences of road scenes taken during the day and during the night. To break the temporal dependence between the frames we only sample every 100-th frame from the day sequences.

\clearpage

\section{Further qualitative results}

\begin{figure*}
  \centering%

\jsubfig{\includegraphics[height=2.8cm]{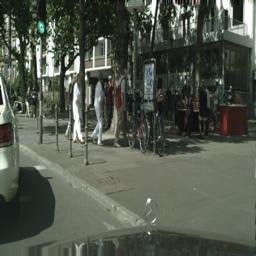}} 
 {}%
\hspace{3pt}
\hfill%
\jsubfig{\includegraphics[height=2.8cm]{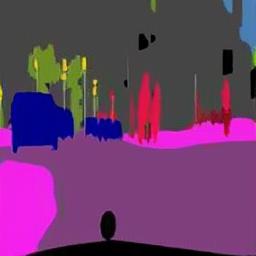}} 
 {}%
\hfill%
\jsubfig{\includegraphics[height=2.8cm]{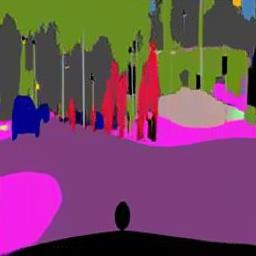}} 
 {}%
\hfill%
\jsubfig{\includegraphics[height=2.8cm]{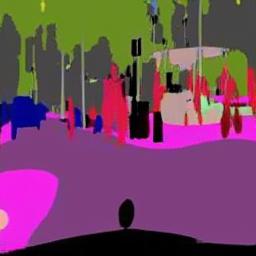}} 
 {}%
\hfill%
\jsubfig{\includegraphics[height=2.8cm]{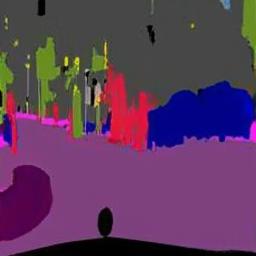}} 
 {}%
\hfill%
\jsubfig{\includegraphics[height=2.8cm]{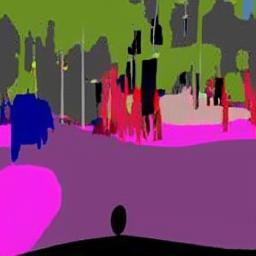}} 
 {}%
\hfill \\ 
\jsubfig{\includegraphics[height=2.8cm]{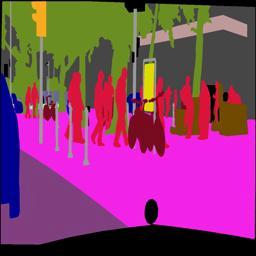}} 
 {}%
\hspace{3pt}
\hfill%
\jsubfig{\includegraphics[height=2.8cm]{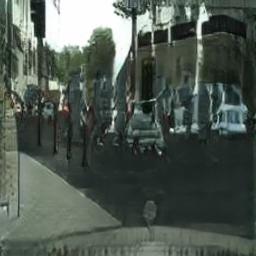}} 
 {}%
\hfill%
\jsubfig{\includegraphics[height=2.8cm]{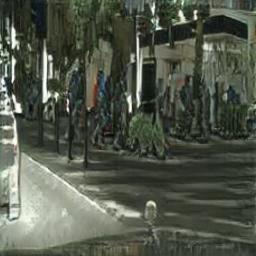}} 
 {}%
\hfill%
\jsubfig{\includegraphics[height=2.8cm]{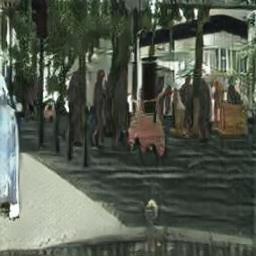}} 
 {}%
\hfill%
\jsubfig{\includegraphics[height=2.8cm]{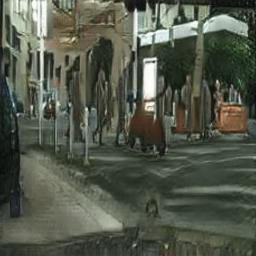}} 
 {}%
\hfill%
\jsubfig{\includegraphics[height=2.8cm]{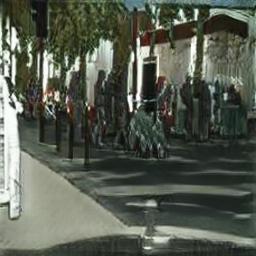}} 
 {}%
\hfill \\

\jsubfig{\includegraphics[height=2.8cm]{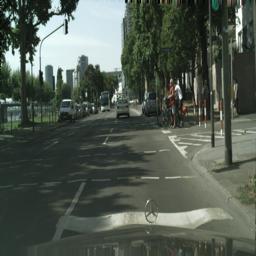}} 
 {}%
\hspace{3pt}
\hfill%
\jsubfig{\includegraphics[height=2.8cm]{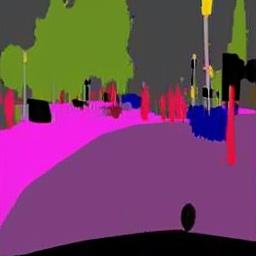}} 
 {}%
\hfill%
\jsubfig{\includegraphics[height=2.8cm]{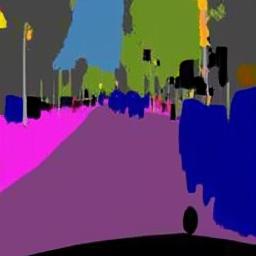}} 
 {}%
\hfill%
\jsubfig{\includegraphics[height=2.8cm]{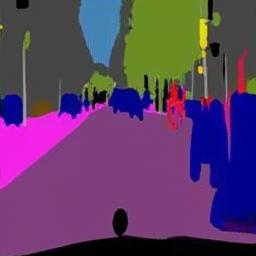}} 
 {}%
\hfill%
\jsubfig{\includegraphics[height=2.8cm]{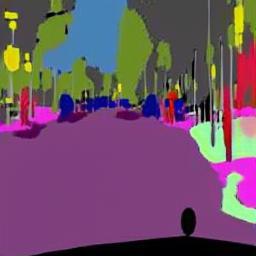}} 
 {}%
\hfill%
\jsubfig{\includegraphics[height=2.8cm]{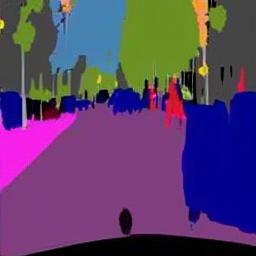}} 
 {}%
\hfill \\ 
\jsubfig{\includegraphics[height=2.8cm]{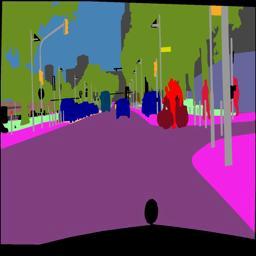}} 
 {}%
\hspace{3pt}
\hfill%
\jsubfig{\includegraphics[height=2.8cm]{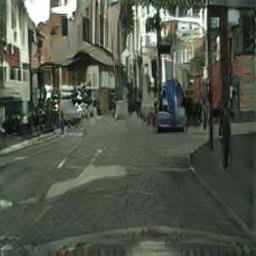}} 
 {}%
\hfill%
\jsubfig{\includegraphics[height=2.8cm]{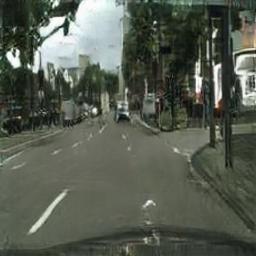}} 
 {}%
\hfill%
\jsubfig{\includegraphics[height=2.8cm]{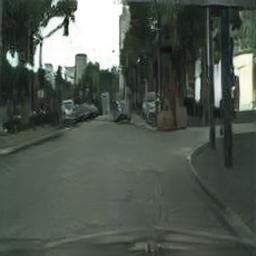}} 
 {}%
\hfill%
\jsubfig{\includegraphics[height=2.8cm]{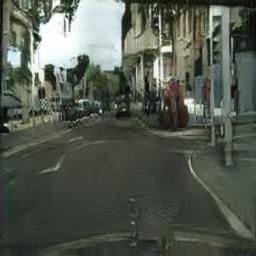}} 
 {}%
\hfill%
\jsubfig{\includegraphics[height=2.8cm]{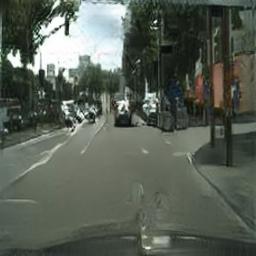}} 
 {}%
\hfill \\

\jsubfig{\includegraphics[height=2.8cm]{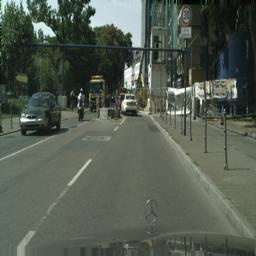}} 
 {}%
\hspace{3pt}
\hfill%
\jsubfig{\includegraphics[height=2.8cm]{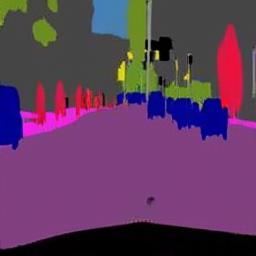}} 
 {}%
\hfill%
\jsubfig{\includegraphics[height=2.8cm]{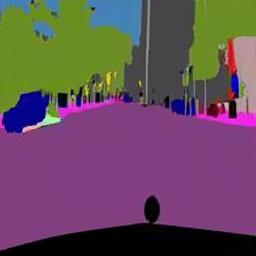}} 
 {}%
\hfill%
\jsubfig{\includegraphics[height=2.8cm]{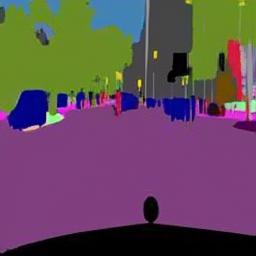}} 
 {}%
\hfill%
\jsubfig{\includegraphics[height=2.8cm]{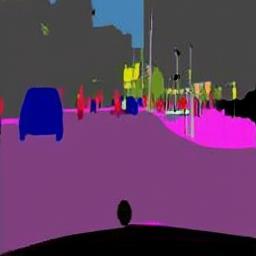}} 
 {}%
\hfill%
\jsubfig{\includegraphics[height=2.8cm]{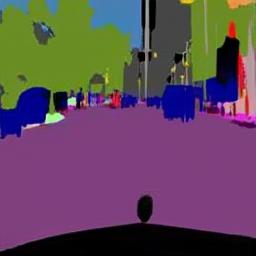}} 
 {}%
\hfill \\ 
\jsubfig{\includegraphics[height=2.8cm]{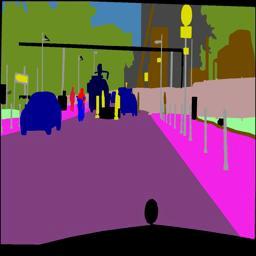}} 
 {Source}%
\hspace{3pt}
\hfill%
\jsubfig{\includegraphics[height=2.8cm]{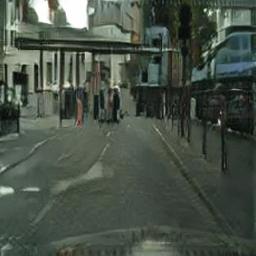}} 
 {$\alpha=0$}%
\hfill%
\jsubfig{\includegraphics[height=2.8cm]{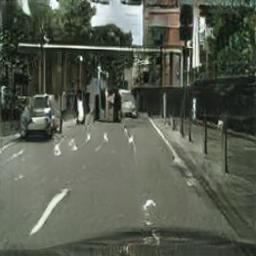}} 
 {$\alpha=0.25$}%
\hfill%
\jsubfig{\includegraphics[height=2.8cm]{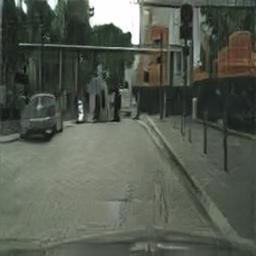}} 
 {$\alpha=0.5$}%
\hfill%
\jsubfig{\includegraphics[height=2.8cm]{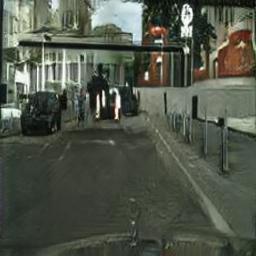}} 
 {$\alpha=0.75$}%
\hfill%
\jsubfig{\includegraphics[height=2.8cm]{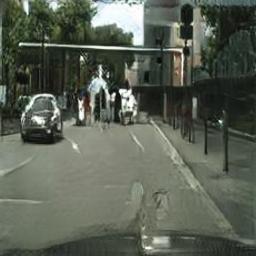}} 
 {$\alpha=1$}%
\hfill \\ 



\vspace{3pt} 
 
 \caption{Additional examples of the effect of learning image-to-image translation with varying pairing ratios } 
 \end{figure*}

\clearpage


\begin{figure*}
\centering%

\jsubfig{\includegraphics[height=3.4cm]{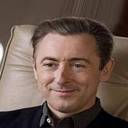}}
	{}%
	\hspace{3pt}
	\jsubfig{\includegraphics[height=3.4cm]{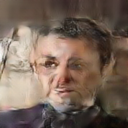}}
	{}%
	\hfill%
	\jsubfig{\includegraphics[height=3.4cm]{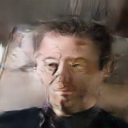}}
	{}%
 	\hfill%
	\jsubfig{\includegraphics[height=3.4cm]{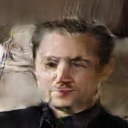}}
	{}%
 	\hfill%
	\hspace{3pt}
	\jsubfig{\includegraphics[height=3.4cm]{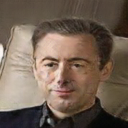}}
	{}%

\jsubfig{\includegraphics[height=3.4cm]{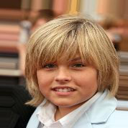}}
	{}%
	\hspace{3pt}
	\jsubfig{\includegraphics[height=3.4cm]{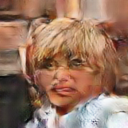}}
	{}%
	\hfill%
	\jsubfig{\includegraphics[height=3.4cm]{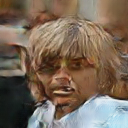}}
	{}%
 	\hfill%
	\jsubfig{\includegraphics[height=3.4cm]{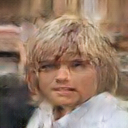}}
	{}%
 	\hfill%
	\hspace{3pt}
	\jsubfig{\includegraphics[height=3.4cm]{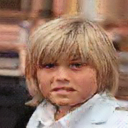}}
	{}%

\jsubfig{\includegraphics[height=3.4cm]{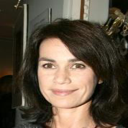}}
	{}%
	\hspace{3pt}
	\jsubfig{\includegraphics[height=3.4cm]{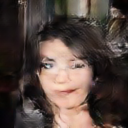}}
	{}%
	\hfill%
	\jsubfig{\includegraphics[height=3.4cm]{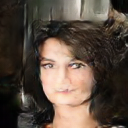}}
	{}%
 	\hfill%
	\jsubfig{\includegraphics[height=3.4cm]{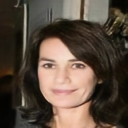}}
	{}%
 	\hfill%
	\hspace{3pt}
	\jsubfig{\includegraphics[height=3.4cm]{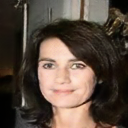}}
	{}%

\jsubfig{\includegraphics[height=3.4cm]{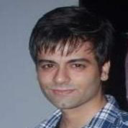}}
	{}%
	\hspace{3pt}
	\jsubfig{\includegraphics[height=3.4cm]{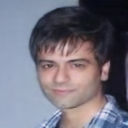}}
	{}%
	\hfill%
	\jsubfig{\includegraphics[height=3.4cm]{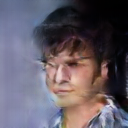}}
	{}%
 	\hfill%
	\jsubfig{\includegraphics[height=3.4cm]{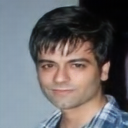}}
	{}%
 	\hfill%
	\hspace{3pt}
	\jsubfig{\includegraphics[height=3.4cm]{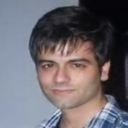}}
	{}%

\jsubfig{\includegraphics[height=3.4cm]{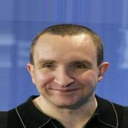}}
	{}%
	\hspace{3pt}
	\jsubfig{\includegraphics[height=3.4cm]{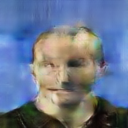}}
	{}%
	\hfill%
	\jsubfig{\includegraphics[height=3.4cm]{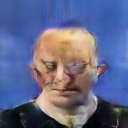}}
	{}%
 	\hfill%
	\jsubfig{\includegraphics[height=3.4cm]{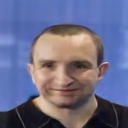}}
	{}%
 	\hfill%
	\hspace{3pt}
	\jsubfig{\includegraphics[height=3.4cm]{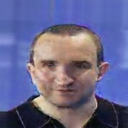}}
	{}%

\jsubfig{\includegraphics[height=3.4cm]{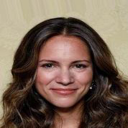}}
{\small{Source}}%
\hspace{3pt}
\jsubfig{\includegraphics[height=3.4cm]{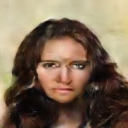}}
{\small{baseline}}%
\hfill%
\jsubfig{\includegraphics[height=3.4cm]{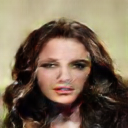}}
{\small{+natural}}%
\hfill%
\jsubfig{\includegraphics[height=3.4cm]{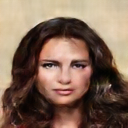}}
{\small{+unpaired}}%
\hfill%
\hspace{3pt}
\jsubfig{\includegraphics[height=3.4cm]{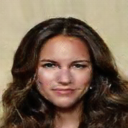}}
{\small{+paired}}%

\vspace{3pt}
\caption{Additional results for the \emph{smile removal} task using different dataset configurations. Above we illustrate our randomly selected results (on the right) compared against three augmentation baselines.
}

\end{figure*}

\begin{figure*}
  \centering%

\jsubfig{\includegraphics[height=3.4cm]{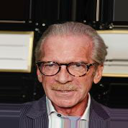}}
	{}%
	\hspace{3pt}
	\jsubfig{\includegraphics[height=3.4cm]{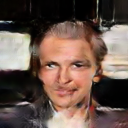}}
	{}%
	\hfill%
	\jsubfig{\includegraphics[height=3.4cm]{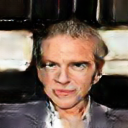}}
	{}%
 	\hfill%
	\jsubfig{\includegraphics[height=3.4cm]{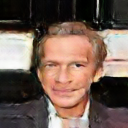}}
	{}%
 	\hfill%
	\hspace{3pt}
	\jsubfig{\includegraphics[height=3.4cm]{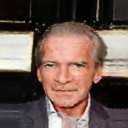}}
	{}%


\jsubfig{\includegraphics[height=3.4cm]{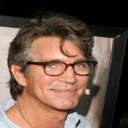}}
	{}%
	\hspace{3pt}
	\jsubfig{\includegraphics[height=3.4cm]{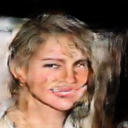}}
	{}%
	\hfill%
	\jsubfig{\includegraphics[height=3.4cm]{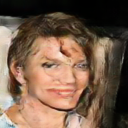}}
	{}%
 	\hfill%
	\jsubfig{\includegraphics[height=3.4cm]{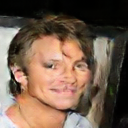}}
	{}%
 	\hfill%
	\hspace{3pt}
	\jsubfig{\includegraphics[height=3.4cm]{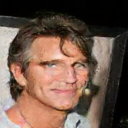}}
	{}%


\jsubfig{\includegraphics[height=3.4cm]{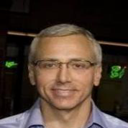}}
	{}%
	\hspace{3pt}
	\jsubfig{\includegraphics[height=3.4cm]{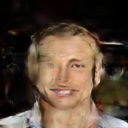}}
	{}%
	\hfill%
	\jsubfig{\includegraphics[height=3.4cm]{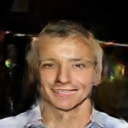}}
	{}%
 	\hfill%
	\jsubfig{\includegraphics[height=3.4cm]{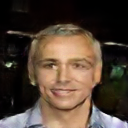}}
	{}%
 	\hfill%
	\hspace{3pt}
	\jsubfig{\includegraphics[height=3.4cm]{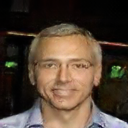}}
	{}%

\jsubfig{\includegraphics[height=3.4cm]{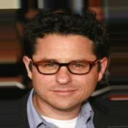}}
	{}%
	\hspace{3pt}
	\jsubfig{\includegraphics[height=3.4cm]{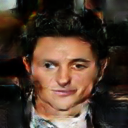}}
    {}%
	\hfill%
	\jsubfig{\includegraphics[height=3.4cm]{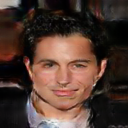}}
	{}%
 	\hfill%
	\jsubfig{\includegraphics[height=3.4cm]{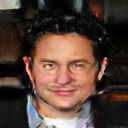}}
	{}%
 	\hfill%
	\hspace{3pt}
	\jsubfig{\includegraphics[height=3.4cm]{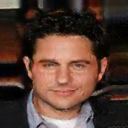}}
	{}%

\jsubfig{\includegraphics[height=3.4cm]{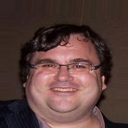}}
 	{}%
	\hspace{3pt}
	\jsubfig{\includegraphics[height=3.4cm]{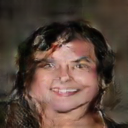}}
 	{}%
	\hfill%
	\jsubfig{\includegraphics[height=3.4cm]{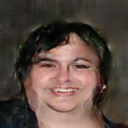}}
 	{}%
 	\hfill%
	\jsubfig{\includegraphics[height=3.4cm]{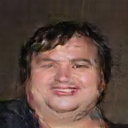}}
 	{}%
 	\hfill%
	\hspace{3pt}
	\jsubfig{\includegraphics[height=3.4cm]{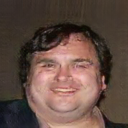}}
 	{}%

\jsubfig{\includegraphics[height=3.4cm]{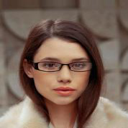}}
 	{\small{Source}}%
	\hspace{3pt}
	\jsubfig{\includegraphics[height=3.4cm]{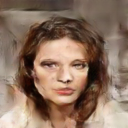}}
 	{\small{baseline}}%
	\hfill%
	\jsubfig{\includegraphics[height=3.4cm]{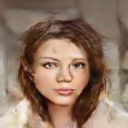}}
 	{\small{+natural}}%
 	\hfill%
	\jsubfig{\includegraphics[height=3.4cm]{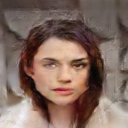}}
 	{\small{+unpaired}}%
 	\hfill%
	\hspace{3pt}
	\jsubfig{\includegraphics[height=3.4cm]{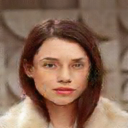}}
 	{\small{+paired}}%

\vspace{3pt}

 \caption{Additional \emph{Eyeglass removal} results using different dataset configurations. Above we illustrate our randomly selected results (on the right) compared against three augmentation baselines.
 } 
 \end{figure*} 



\begin{figure*}
  \centering%

\jsubfig{\includegraphics[height=3.4cm]{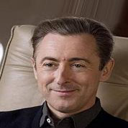}}
	{}%
	\hspace{3pt}
	\jsubfig{\includegraphics[height=3.4cm]{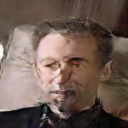}}
	{}%
	\hfill%
	\jsubfig{\includegraphics[height=3.4cm]{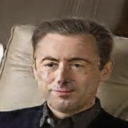}}
	{}%
 	\hfill%
	\jsubfig{\includegraphics[height=3.4cm]{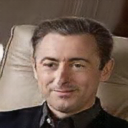}}
	{}%
 	\hfill%
	\jsubfig{\includegraphics[height=3.4cm]{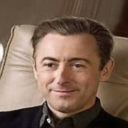}}
	{}%

\jsubfig{\includegraphics[height=3.4cm]{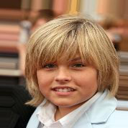}}
	{}%
	\hspace{3pt}
	\jsubfig{\includegraphics[height=3.4cm]{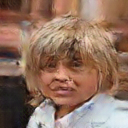}}
	{}%
	\hfill%
	\jsubfig{\includegraphics[height=3.4cm]{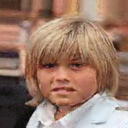}}
	{}%
 	\hfill%
	\jsubfig{\includegraphics[height=3.4cm]{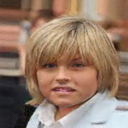}}
	{}%
 	\hfill%
	\jsubfig{\includegraphics[height=3.4cm]{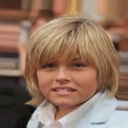}}
	{}%

\jsubfig{\includegraphics[height=3.4cm]{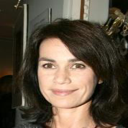}}
	{}%
	\hspace{3pt}
	\jsubfig{\includegraphics[height=3.4cm]{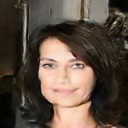}}
	{}%
	\hfill%
	\jsubfig{\includegraphics[height=3.4cm]{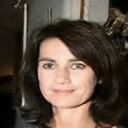}}
	{}%
 	\hfill%
	\jsubfig{\includegraphics[height=3.4cm]{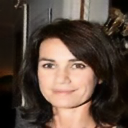}}
	{}%
 	\hfill%
	\jsubfig{\includegraphics[height=3.4cm]{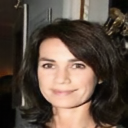}}
	{}%

\jsubfig{\includegraphics[height=3.4cm]{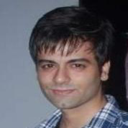}}
	{}%
	\hspace{3pt}
	\jsubfig{\includegraphics[height=3.4cm]{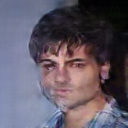}}
	{}%
	\hfill%
	\jsubfig{\includegraphics[height=3.4cm]{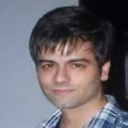}}
	{}%
 	\hfill%
	\jsubfig{\includegraphics[height=3.4cm]{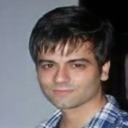}}
	{}%
 	\hfill%
	\jsubfig{\includegraphics[height=3.4cm]{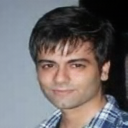}}
	{}%

\jsubfig{\includegraphics[height=3.4cm]{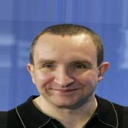}}
	{}%
	\hspace{3pt}
	\jsubfig{\includegraphics[height=3.4cm]{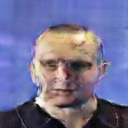}}
	{}%
	\hfill%
	\jsubfig{\includegraphics[height=3.4cm]{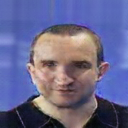}}
	{}%
 	\hfill%
	\jsubfig{\includegraphics[height=3.4cm]{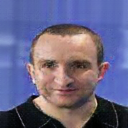}}
	{}%
 	\hfill%
	\jsubfig{\includegraphics[height=3.4cm]{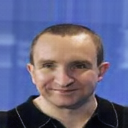}}
	{}%

\jsubfig{\includegraphics[height=3.4cm]{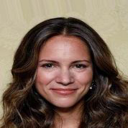}}
{\small{Source}}%
\hspace{3pt}
\jsubfig{\includegraphics[height=3.4cm]{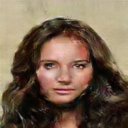}}
{\small{25\%}}%
\hfill%
\jsubfig{\includegraphics[height=3.4cm]{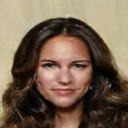}}
{\small{50\%}}%
\hfill%
\jsubfig{\includegraphics[height=3.4cm]{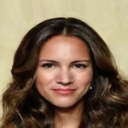}}
{\small{75\%}}%
\hfill%
\jsubfig{\includegraphics[height=3.4cm]{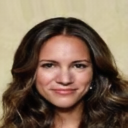}}
{\small{100\%}}%

 \caption{Additional pseudo-pair ratio analysis results for the  \emph{smile removal} task}
 \end{figure*} 


\begin{figure*}
  \centering%

\jsubfig{\includegraphics[height=3.4cm]{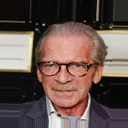}}
	{}%
	\hspace{3pt}
	\jsubfig{\includegraphics[height=3.4cm]{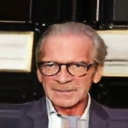}}
	{}%
	\hfill%
	\jsubfig{\includegraphics[height=3.4cm]{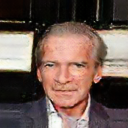}}
	{}%
 	\hfill%
	\jsubfig{\includegraphics[height=3.4cm]{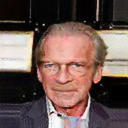}}
	{}%
 	\hfill%
	\jsubfig{\includegraphics[height=3.4cm]{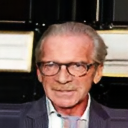}}
	{}%


\jsubfig{\includegraphics[height=3.4cm]{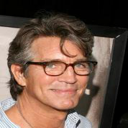}}
	{}%
	\hspace{3pt}
	\jsubfig{\includegraphics[height=3.4cm]{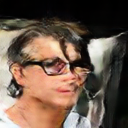}}
	{}%
	\hfill%
	\jsubfig{\includegraphics[height=3.4cm]{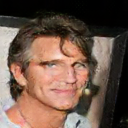}}
	{}%
 	\hfill%
	\jsubfig{\includegraphics[height=3.4cm]{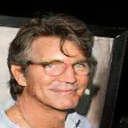}}
	{}%
 	\hfill%
	\jsubfig{\includegraphics[height=3.4cm]{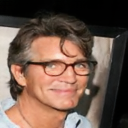}}
	{}%
	

\jsubfig{\includegraphics[height=3.4cm]{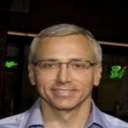}}
	{}%
	\hspace{3pt}
	\jsubfig{\includegraphics[height=3.4cm]{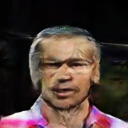}}
	{}%
	\hfill%
	\jsubfig{\includegraphics[height=3.4cm]{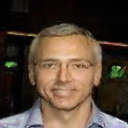}}
	{}%
 	\hfill%
	\jsubfig{\includegraphics[height=3.4cm]{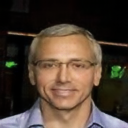}}
	{}%
 	\hfill%
	\jsubfig{\includegraphics[height=3.4cm]{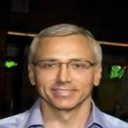}}
	{}%

\jsubfig{\includegraphics[height=3.4cm]{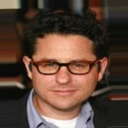}}
	{}%
	\hspace{3pt}
	\jsubfig{\includegraphics[height=3.4cm]{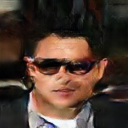}}
	{}%
	\hfill%
	\jsubfig{\includegraphics[height=3.4cm]{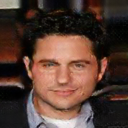}}
    {}%
 	\hfill%
	\jsubfig{\includegraphics[height=3.4cm]{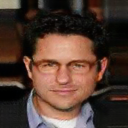}}
	{}%
 	\hfill%
	\jsubfig{\includegraphics[height=3.4cm]{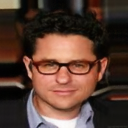}}
	{}%

\jsubfig{\includegraphics[height=3.4cm]{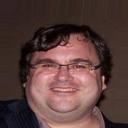}}
	{}%
	\hspace{3pt}
	\jsubfig{\includegraphics[height=3.4cm]{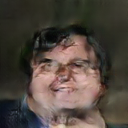}}
	{}%
	\hfill%
	\jsubfig{\includegraphics[height=3.4cm]{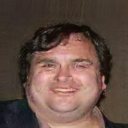}}
	{}%
 	\hfill%
	\jsubfig{\includegraphics[height=3.4cm]{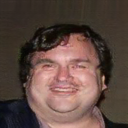}}
	{}%
 	\hfill%
	\jsubfig{\includegraphics[height=3.4cm]{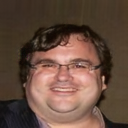}}
	{}%

\jsubfig{\includegraphics[height=3.4cm]{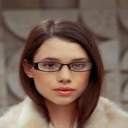}}
	{\small{Source}}%
	\hspace{3pt}
	\jsubfig{\includegraphics[height=3.4cm]{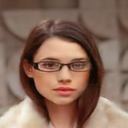}}
	{\small{25\%}}%
	\hfill%
	\jsubfig{\includegraphics[height=3.4cm]{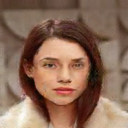}}
	{\small{50\%}}%
 	\hfill%
	\jsubfig{\includegraphics[height=3.4cm]{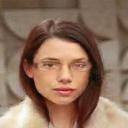}}
	{\small{75\%}}%
 	\hfill%
	\jsubfig{\includegraphics[height=3.4cm]{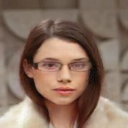}}
	{\small{100}}%

\vspace{3pt} 
 
 \caption{Additional pseudo-pair ratio analysis results for the  \emph{eyeglass removal} task} 
 \end{figure*} 


\begin{figure*}
  \centering%

\jsubfig{\includegraphics[height=3.4cm]{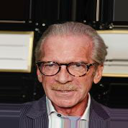}}
	{}%
	\hspace{3pt}
	\jsubfig{\includegraphics[height=3.4cm]{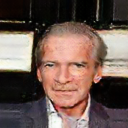}}
	{}%
	\hspace{1pt}
	\jsubfig{\includegraphics[height=3.4cm]{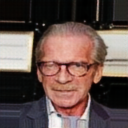}}
	{}%


\jsubfig{\includegraphics[height=3.4cm]{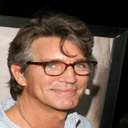}}
	{}%
	\hspace{3pt}
	\jsubfig{\includegraphics[height=3.4cm]{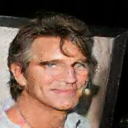}}
	{}%
	\hspace{1pt}
	\jsubfig{\includegraphics[height=3.4cm]{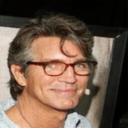}}
	{}%


\jsubfig{\includegraphics[height=3.4cm]{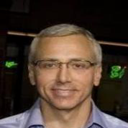}}
	{}%
	\hspace{3pt}
	\jsubfig{\includegraphics[height=3.4cm]{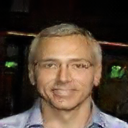}}
	{}%
	\hspace{1pt}
	\jsubfig{\includegraphics[height=3.4cm]{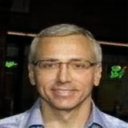}}
	{}%

\jsubfig{\includegraphics[height=3.4cm]{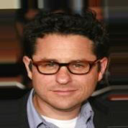}}
	{}%
	\hspace{3pt}
	\jsubfig{\includegraphics[height=3.4cm]{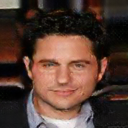}}
	{}%
	\hspace{1pt}
	\jsubfig{\includegraphics[height=3.4cm]{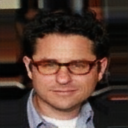}}
	{}%

\jsubfig{\includegraphics[height=3.4cm]{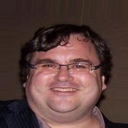}}
 	{}%
	\hspace{3pt}
	\jsubfig{\includegraphics[height=3.4cm]{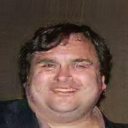}}
 	{}%
	\hspace{1pt}
	\jsubfig{\includegraphics[height=3.4cm]{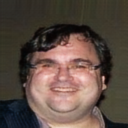}}
 	{}%

\jsubfig{\includegraphics[height=3.4cm]{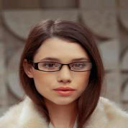}}
	{\small{Source}}%
	\hspace{3pt}
	\jsubfig{\includegraphics[height=3.4cm]{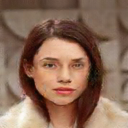}}
	{\small{Implicit}}%
	\hspace{1pt}
	\jsubfig{\includegraphics[height=3.4cm]{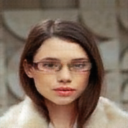}}
	{\small{Explicit}}%
	
	
%


 \caption{Additional \emph{Eyeglass removal} results with pseudo-pairs in different settings.} 
 \end{figure*} 



\end{document}